\documentclass[journal]{IEEEtran}
\usepackage{url}
\usepackage{amsmath}
\usepackage{graphicx, color, booktabs, multirow, amssymb}
\usepackage{subfigure}
\usepackage{float}

\usepackage{comment}
\usepackage[linesnumbered, ruled, vlined]{algorithm2e}
\makeatletter

\floatstyle{ruled}
\newfloat{algorithm}{tbp}{loa}
\providecommand{\algorithmname}{Algorithm}
\floatname{algorithm}{\protect\algorithmname}

\usepackage{algpseudocode}
\algrenewcommand{\algorithmicrequire}{\textbf{Input: }}
\algrenewcommand{\algorithmicensure}{\textbf{Output: }}
\algrenewcommand{\algorithmicforall}{\textbf{for each}}

\newcommand{\revised}[1]{{\color{black} #1}}

\makeatother

\begin{document}
\title{Tiny Adversarial Multi-Objective Oneshot Neural Architecture Search}
\author{Guoyang~Xie,
        Jinbao~Wang,
        Guo~Yu,~\IEEEmembership{Member,~IEEE,}
        Feng~Zheng,~\IEEEmembership{Member,~IEEE,}
        and~Yaochu~Jin,~\IEEEmembership{Fellow,~IEEE}
\thanks{G. Xie is with the Department of Computer Science, University of Surrey, Guildford, Surrey GU2 7XH, U.K. He is also with the Department of Computer Science, Southern University of Science and Technology, Shenzhen 518055, China. (e-mail: guoyang.xie@surrey.ac.uk)}
\thanks{J. Wang is with the Department of Computer Science, Southern University of Science and Technology, Shenzhen, 518055, China (e-mail: linkingring@163.com) }
\thanks{G. Yu is with the Key Laboratory of Advanced Control and Optimization for Chemical Processes, Ministry of Education, East China University of Science and Technology, Shanghai 200237, China (e-mail: guoyu@ecust.edu.cn) }
\thanks{F. Zheng is with the Department of Computer Science, Southern University of Science and Technology, Shenzhen, 518055, China (e-mail: zhengf@sustech.edu.cn) }
\thanks{Y. Jin is with the Chair of Nature Inspired Computing and Engineering, Faculty of Technology, Bielefeld University, D-33615 Bielefleld, Germany. He is also with the Department of Computer Science, University of Surrey, Guildford, GU2 7XH, U.K.  (e-mail: yaochu.jin@surrey.ac.uk) (\textit{corresponding authors: Yaochu Jin, Feng Zheng})}
}

\maketitle

\begin{abstract}
Due to limited computational cost and energy consumption, most neural network models deployed in mobile devices are tiny. However, tiny neural networks are commonly very vulnerable to attacks. Current research has proved that larger model size can improve robustness, but little research focuses on how to enhance the robustness of tiny neural networks. Our work focuses on how to improve the robustness of tiny neural networks without seriously deteriorating of clean accuracy under mobile-level resources. To this end, we propose a multi-objective oneshot network architecture search (NAS) algorithm to obtain the best trade-off networks in terms of the adversarial accuracy, the clean accuracy and the model size. Specifically, we design a novel search space based on new tiny blocks and channels to balance model size and adversarial performance. Moreover, since the supernet significantly affects the performance of subnets in our NAS algorithm, we reveal the insights into how the supernet helps to obtain the best subnet under white-box adversarial attacks. Concretely, we explore a new adversarial training paradigm by analyzing the adversarial transferability and the difference between training the subnets from scratch and fine-tuning. Finally, we make a statistical analysis for the layer-wise combination of certain blocks and channels on the first non-dominated front, which can serve as a guideline to design tiny neural network architectures for the resilience of adversarial perturbations.
\end{abstract}


\IEEEpeerreviewmaketitle

\section{Introduction}

\IEEEPARstart{I}{t} is well known that deep neural networks are vulnerable to attacks that add small perturbations to the input data, which are almost imperceptible to human vision systems \cite{szegedy:szegedy2013intriguing, goodfellow:goodfellow2014explaining}. The maliciously perturbed examples are commonly obtained by two operations. One is to add pixel-wise $\epsilon-$bounded perturbations to the input data \cite{madry:madry2017towards}. The other operation is to generate examples using the unrestricted perturbation like rotation, spatially translations
\cite{xiao:xiao2018spatially}. In this work, we focus on defensive mechanisms against the former, although they can be extended to the latter.

\begin{figure}[t]
    \centering
    \includegraphics[width=0.5\textwidth]{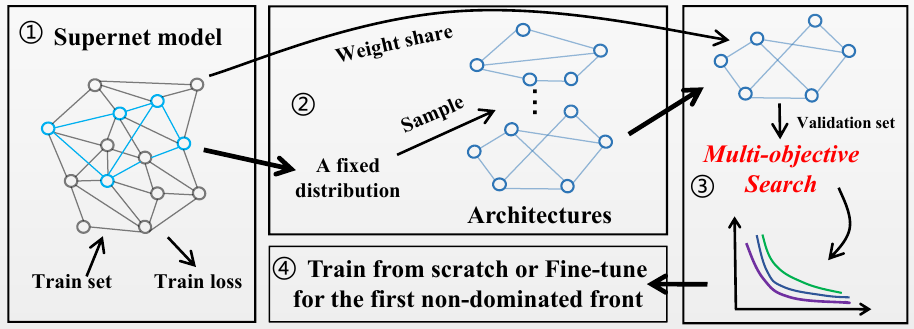}
    \caption{Multi-objective oneshot NAS framework. The first step is to design the supernet search space and the loss function. The second step is to sample different subnets architectures from the supernet under a fixed distribution and train the subnets for a small number of epochs. The third step is to multi-objective search the best trade-off subnets and obtain the fitness value of the subnets by cloning the weight from the sueprnet. The fourth step is to train the non-dominated subnets.}
    \label{fig:easy_framework}
\end{figure}

To defend the attacks, current research \cite{madry:madry2017towards, cubuk:cubuk2017intriguing, afs} constantly adopt ResNet \cite{he:he2016deep} as the backbone network to \revised{explore the relationship between the capacity of the model and its adversarial robustness.} \revised{It has been empirically proved that increasing the number of parameters of neural networks is able to improve robustness}. However, little research pays attention to the improvement on robustness of tiny neural networks, which are widely used for mobile applications. Their sizes typically range from 10K to 2M. Therefore, our work focuses on the balance between the adversarial defensive ability and tiny model size.

The trade-off between the adversarial accuracy and the clean accuracy have been examined in \cite{tsipras:tsipras2018robustness}. The work in \cite{tsipras:tsipras2018robustness} uses one specific binary classifier to illustrate the trade-off, which can be easily generalized to multi-class problems with different assumptions \cite{zhang:zhang2019theoretically}. Previous work have illustrated the importance of the network architecture for adversarial robustness \cite{madry:madry2017towards, cubuk:cubuk2017intriguing}. Most of them add a specific layer to the existing architecture to increase the adversarial accuracy, which, however, do not take into account the potential relationship between different layers in the whole neural network architecture. Our aim is to find the best trade-off neural networks with respect to the clean accuracy, the adversarial accuracy and tiny model size. Hence, we need to take a global view to redesign tiny neural networks. Our assumption is that designing a tiny robust neural network without the loss of clean accuracy can be transformed into a combinational optimization problem of different layers with what kinds of blocks and channels. Hence, we propose a tiny adversarial multi-objective oneshot network architecture search (TAM-NAS) to search the best trade-off solutions.   

There are different defensive training strategies for adversarial attacks. The most common approach \cite{goodfellow:goodfellow2014explaining} is to train neural networks on adversarial examples. Recently, another popular approach is to formulate adversarial training as a min-max robust optimization problem \cite{madry:madry2017towards, wong:wong2018scaling}. The inner maximization problem is to find the worst performance in the presence of perturbations in the input data and the outer minimization problem is to find the optimal model parameters given the worst-case perturbation. The state-of-the-art defensive result is reported in \cite{zhang:zhang2019theoretically}, which proposes a new classification-calibrated surrogate loss function to simultaneously minimize the clean error and the boundary error. The boundary error refers to the neural network output differences between the original input data and adversarial examples. But the major issue of adversarial training is the high computational cost for adversarial training. To speed up our adversarial training for each epoch, we use TRADES-YOPO-m-n from \cite{zhang:zhang2019you} as our adversarial training method. TRADES-YOPO-m-n aims not only to incorporate the surrogate loss function from \cite{zhang:zhang2019theoretically} but also to reduce the computational cost by restricting most of forward and backward propagations within the first layer of the network during adversary training. As a result, we are able to reduce the training time to 2.5 minutes for each epoch on a single V100 GPU platform.   

\begin{figure*}[htb]
    \centering
    \includegraphics[width=1\textwidth]{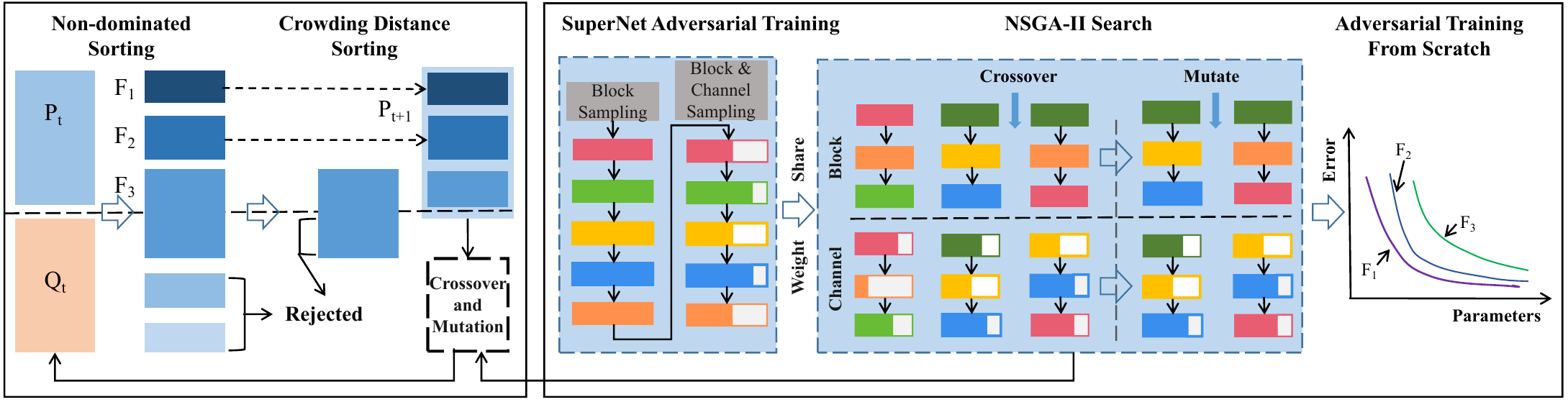}
    \caption{The overall framework of TAM-NAS. The first step is to design a supernet search space and uniform sample the new subnet candidates. Our sampling strategy is divided into two phase. One is the block sampling phase, the other is block and channel jointly sampling phase. The second step is to adversarial train the subnets sampled from the supernet. The third step is to multi-objective search the best trade-off subnets. The fitness value of the subnets is achieved by cloning the weight from the supernet. The final step is to train from scratch or fine-tune the non-dominated subnets. }
    \label{fig:framework}
\end{figure*}

Human experts make tons of effort to devise several classical architectures, such as ResNet \cite{he:he2016deep} and Inception \cite{szegedy:szegedy2015going}. But it is not feasible to manually explore an infinite number of architectures. As an emerging automated machine learning technique, neural network architecture search (NAS) \cite{liu:liu2018darts, pham:pham2018efficient} is prevalent because it requires much less expertise and effort to discover good architectures for a new task and dataset. Most NAS approaches employ reinforcement learning \cite{zoph:zoph2018learning}, evolutionary algorithms \cite{xie:xie2017genetic}, and gradient-based methods \cite{liu:liu2018darts} to design neural networks automatically. However, most of them treat it as a single-objective optimization problem, which is not well suited for solving the trade-off problem. Inspired by the work in \cite{jin2008pareto, MRL, modulenet}, we employ a multi-objective approach based NAS to find the best architecture for the trade-off solutions between the adversarial accuracy, the clean accuracy and the mode size. In our work, we mainly address the trade-off problem based on the ShuffleNetV2 architecture \cite{ma2018shufflenet}, Xception block \cite{chollet2017xception}, SE layer \cite{hu2018squeeze},  Non-Local block \cite{wang2018non}, and their variants.

\textbf{Our contributions can be summarized as follows}:

\begin{itemize}
\item To find the best trade-off neural networks between the adversarial accuracy, the clean accuracy and the model size, we specifically propose three novel tiny robust blocks. Due to the inertial self-attention mechanism, the layer-wise combination of these three blocks can increase robustness without significantly worsening clean performance.
\item We explore a new adversarial training paradigm for the supernet. Because the subnets highly rely on the supernet in oneshot NAS, the adversarial performance of the subnets could be further improved by using our proposed training paradigm. To this end, we analyse how the width of supernet, the perturbation range and the number of attack steps for the supernet adversarial training affect the performance of the subnets. 
\item We seamlessly integrate multi-objective search algorithm with one-shot NAS algorithm. After finishing the search procedure, we can directly obtain the non-dominated front which can speed up finding the best trade-off subnets. In addition, we discover that training from scratch outperforms fine-tuning for the non-dominated subnets.
\item We draw a conclusion about how to design a tiny robust neural network. Firstly, the robust blocks, i.e. pure robust blocks and tiny robust blocks should be put into the shallow layers while pure tiny blocks should be put into the deep layers. Secondly, larger intermediate channels should be put into the shallow layers and the intermediate channels should gradually decline in the rest of layers. Finally, we rebuild a tiny neural network according to the provided guideline and find that it can not only reduces the model size but also increases the adversarial accuracy and clean accuracy. 
\end{itemize}

\section{Related Work}
\textbf{Adversarial Training} is the most common defensive mechanism against adversarial attacks \cite{goodfellow:goodfellow2014explaining}, which uses both clean and adversarial images for training. Originated from game theory \cite{myerson2013game}, the work in \cite{grosshans2013bayesian} reformulates the min-max optimization problem of adversarial learning as Nash equilibrium \cite{daskalakis2009complexity}. The game-theory based optimization method \cite{zhang:zhang2019you} can effectively reduce the high computational cost without sacrificing adversarial accuracy. Most of existing approaches only focus on improving robustness but ignore the deterioration of clean accuracy. Especially, the work \cite{tsipras:tsipras2018robustness} has empirically proved the trade-off information between the adversarial accuracy and the clean accuracy. Kannan et al. \cite{kannan2018adversarial} and Zhang et al. \cite{zhang:zhang2019theoretically} construct the surrogate loss function to make the difference of clean images and adversarial counterparts smaller. Madry et al. \cite{madry:madry2017towards} conclude that larger capacity size of neural networks can improve the performance under adversarial attacks. it is known that most neural network models running in our electronic devices are very tiny because the devices have constraints on the energy consumption and the amount of storage for the models. Hence, we aim to figure out which kind of tiny neural network architectures can be effective for the resilience of adversarial perturbations. 

\textbf{Neural Network Architecture Search} aims to replace handcrafted architecture search with automated machine learning technique. Representative search algorithms includes evolutionary algorithms \cite{xie:xie2017genetic, real2019regularized}, reinforcement learning \cite{zoph:zoph2018learning, pham:pham2018efficient}, and gradient-based methods \cite{liu:liu2018darts, dong:dong2019one,guo2020meets}. In one-shot NAS \cite{bender2018understanding}, the authors \cite{bender2018understanding} construct a supernet which can generate every possible architecture in the search space. The work in \cite{bender2018understanding} train a supernet for once and then at search time, they can obtain various fitness value for different subnets by weight sharing from the supernet. But most of them employ a single-objective optimization approach to search, which is not well suited for solving the trade-off problem \cite{tsipras:tsipras2018robustness}. In order to solve the multi-objective optimization problem, we adopt the elitist non-dominated sorting genetic algorithm (NSGA-II) \cite{deb2002fast} as our search algorithm. Inspired by \cite{xie2019intriguing}, we try to investigate the influence of neural network depth on network resilience as the adversarial attack. Furthermore, we also investigate the relationship between the supernet and its subnets in our one-shot multi-objective NAS framework and give a hint on how to adversarial train supernet to obtain the better adversarial performance of subnets. 

\section{Tiny Adversarial Multi-Objective One-Shot NAS}

Our NAS approach consists of four steps as following. (1) Design a supernet search space and uniformly sample different candidates from supernet to increase our supernet representation ability for a number of subnet architectures when using a single supernet. (2) Train the candidates sampled from the supernet on adversarial examples and make it more robust in the presence of the adversarial attacks. (3) Multi-objective search the new subnet by using the elitist non-dominated sorting generic algorithm (NSGA-II) \cite{deb2002fast} and evaluate the clean accuracy, the adversarial accuracy, and the number of parameters of each subnet by cloning it weight from the pre-trained supernet. (4) Fine-tune each subnet from the first non-dominated front and evaluate their performance on the test dataset. Fig. \ref{fig:framework} shows our overall framework.

\subsection{\textbf{Problem Definition}}
Without loss of generality, our supernet search space $\mathcal{A}$ can be represented by a directed acyclic graph (DAG), denoted as $\mathcal{N}(\mathcal{A}, W)$, where $W$ is the weight of supernet. A subnet architecture is a subgraph $a \in \mathcal{A}$, denoted as $\mathcal{N}(a, w)$, where $w$ is the weight of subnet. $\Gamma(\mathcal{A})$ is a prior distribution of $a \in \mathcal{A}$. $\mathcal{L}_{adv-train}\left ( \cdot  \right )$ is the adversarial training loss function on the adversarial training examples. The most important factor for TAM-NAS is that the performance of the subnets using inherited weights from supernet (without extra fine-tuning or training from scratch) should be highly predictive. In other words, the supernet weights $W_{\mathcal{A}}$ should be optimized in a way that all subnet architectures in the search space $A$ are optimized simultaneously. It can be expressed as in Eq. (1), 
\begin{equation} \label{eq:supernet_definition}
    W_{\mathcal{A}} = \underset{W}{\mathrm{argmin}} \;\mathbb{E}_{a \sim \Gamma(\mathcal{A})}\left [\mathcal{L}_{adv-train}(\mathcal{N}(a, W(a)))  \right ].
\end{equation}
After finishing the training of supernet, the next step is to find a set of Pareto optimal subnets $a^{*} \in \mathcal{A}$ in terms of our objectives: the adversarial error, the clean error, and the model size. It can be expressed as in Eq. (2),

\begin{equation}\label{eq:multi_objective_definition}
\begin{aligned}
    \min& \left \{f_{1}(a^{*}), f_{2}(a^{*}), f_{3}(a^{*})\right \}\\
    \textrm{s.t.}&\; a^{*} \in \mathcal{A}
\end{aligned}
\end{equation}
\revised{where $f_{1}, f_{2}, f_{3}$ are the three objectives, the adversarial error, the clean error and the model size, namely. Actually, it is not able to get the minimum value for three objectives simultaneously since there is a strong trade-off relationship between each pair of the objectives. For instance, if the model size is larger, the adversarial error and the clean error will become smaller. Our aim is to obtain a tiny model with compatible performance in adversarial dataset and clean dataset. Fig. \ref{fig:easy_framework} shows the pipeline of multi-objective oneshot NAS.}

\subsection{\textbf{Search Space Design}}
Since we aim to search tiny robust neural networks, our supernet \revised{adopts} one of state-of-the-art hand-crafted tiny network architecture--ShuffleNetV2 \cite{ma2018shufflenet} as the backbone model. Since our experiments are mainly conducted on the CIFAR10 \cite{krizhevsky2009learning} and SVHN \cite{netzer2011reading} datasets, the depth and width of the supernet are largely different from the original ShuffleNetV2 which is developed on Imagenet dataset \cite{ILSVRC15}. Table. \ref{tab:supernet_architecture} shows the parameter setting of the overall architecture of the supernet. BN represents the batch norm layer. $3\times3$ Conv represents a convolutional layer and its kernel size is 3. CB refers to the choice block chosen from our predefined block search space. SE refers to the SE layer \cite{hu2018squeeze}. Moreover, we design search space for the channel number search of each choice block. In total, we provide 22 block choices and 10 channel number choices for the search space. Below, we will separately describe our search space in detail. 
\begin{table}[ht]
    \centering
    \caption{Supernet Architecture}
    \begin{tabular}{c|c|c|c|c}
        \hline
        Input Shape & Block & Channels & Repeat & Stride  \\
        \hline
        $32^{2}\times$3 & 3$\times$3 Conv & 24 & 1 & 1 \\
        \hline
        $32^{2}\times$24 & BN & 24 & 1 &  \\
        \hline
        $32^{2}\times$24 & CB & 48 & 1 & 2 \\
        \hline
        $16^{2}\times$48 & CB & 48 & 3 & 1 \\
        \hline
        ${16}^{2}\times$48 & CB & 96 & 1 & 2 \\
        \hline
        ${8}^{2}\times$96 & CB & 96 & 7 & 1 \\
        \hline
        ${8}^{2}\times$96 & CB & 192 & 1 & 2 \\
        \hline
        ${4}^{2}\times$192 & CB & 192 & 3 & 1 \\
        \hline
        ${4}^{2}\times$192 & 1$\times$1 Conv & 176 & 1 & 1 \\
        \hline
        ${4}^{2}\times$176 & BN & 176 & 1 \\
        \hline
        ${1}^{2}\times$176 & Pooling & 176 & 1 \\
        \hline
        ${1}^{2}\times$176 & SE & 920 & 1 \\
        \hline
        ${1}^{2}\times$920 & 1$\times$1 Conv & 1024 & 1 & 1 \\
        \hline
        ${1}^{2}\times$1024 & FC & 10 & 1 \\
        \hline
    \end{tabular}
    \label{tab:supernet_architecture}
\end{table}

Three kinds of blocks we used in block search spaces as followings.

\subsubsection{\textbf{Pure Tiny Blocks}}
\revised{Pure tiny blocks mainly come from ShuffleNetV2 \cite{ma2018shufflenet}. We add the self-attention layer--SE layer \cite{hu2018squeeze} into the main branch for balancing between the accuracy and inference speed. Fig. \ref{fig:shuffl2v2_block_choices} (a) and Fig. \ref{fig:shuffl2v2_block_choices} (b) are the pure tiny blocks when the non-local layers from the main branch are removed.}

\subsubsection{\textbf{Pure Robust Blocks}}
Firstly, we design a non-local block for image denoising, inspired by \cite{wang2018non}. We also add another self-attention layer--SE layer \cite{hu2018squeeze} as the last layer of the non-local block since it has been found \cite{wang2018non} that the self-attention mechanism could make neural network more robust. Fig. \ref{fig:shuffl2v2_block_choices} shows two non-local blocks which refer to as Embedded Gaussian version and Gaussian version \cite{wang2018non}. Fig. \ref{fig:shuffl2v2_block_choices} (b) shows the internal architecture of the pure robust blocks.

\subsubsection{\textbf{Tiny Robust Blocks}}
\begin{figure*}
\begin{center}
\includegraphics[width=0.95\linewidth]{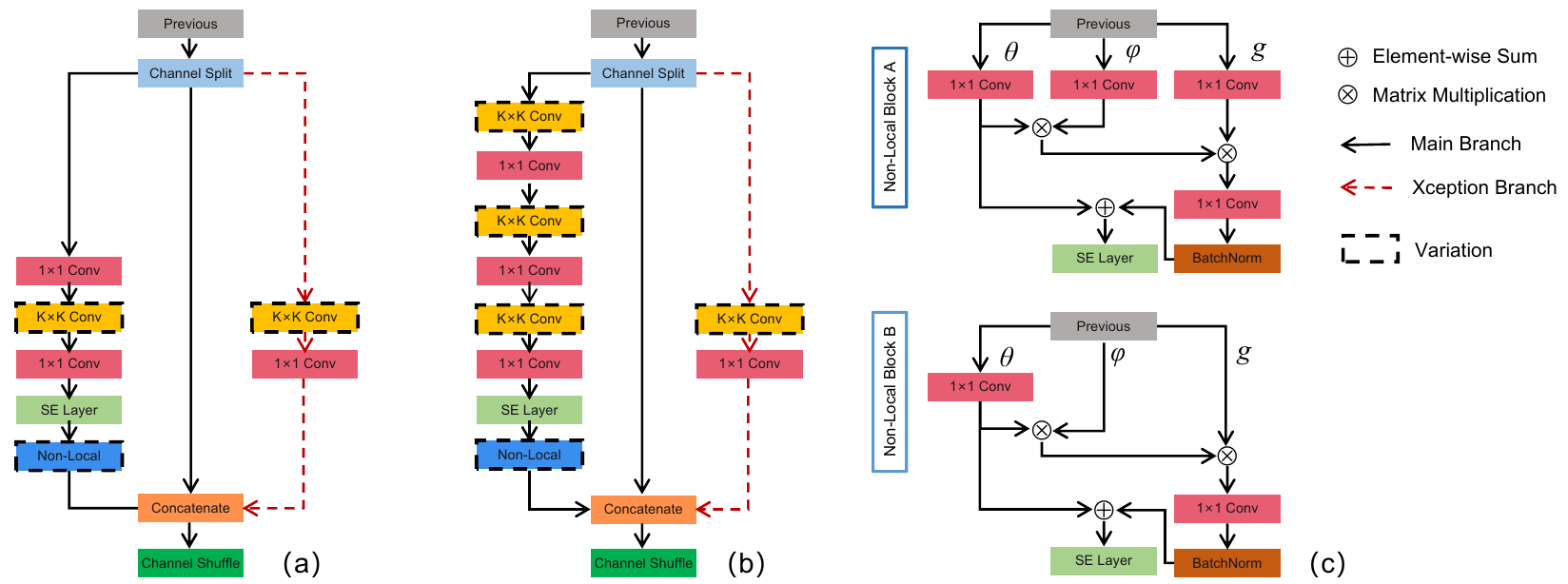}
\end{center}
\caption{(a) and (b) represent the internal architectures of the tiny robust blocks. K refer to the kernel size and it ranges among 3, 5, 7. The dashed line indicates the internal search space for the tiny robust blocks. So if we removed the non-local layer from the main branch of (a) and (b), (a) and (b) will denote as the pure tiny blocks. (c) represents the internal architectures of the pure robust blocks. The upper part of (c) represents the combination of SE layer and Embedded-Gaussian non-local layer. The bottom part of (c) represents the combination of SE layer and Gaussian non-local layer.}
\label{fig:shuffl2v2_block_choices}
\end{figure*}

To make pure tiny blocks become more robust, we try to add the non-local block and the SE layer into the main branch of the original shufflev2 and shufflev2-xception block. Figs. \ref{fig:shuffl2v2_block_choices} (a) and (b) show the internal architecture of the tiny robust block. Their kernel sizes of depth-wise convolutional layer range among 3, 5, 7. Furthermore, since the non-local layer will add more parameters for our shufflev2 or shufflev2-xception block, we set the non-local layer of shufflev2 and shufflev2-xception block as an optional choice, which means it belongs to another new search space. When we remove the non-local layer and SE layer in tiny robust blocks, they will work as pure tiny blocks. In Fig. \ref{fig:shuffl2v2_block_choices}, the dashed line indicates the internal search space for each block. The encoding method of choice blocks is to assign an order from 0 to 21 for each block.

\subsubsection{\textbf{Channel Search Spaces}}
\begin{figure*}
\begin{center}
\includegraphics[width=0.95\linewidth]{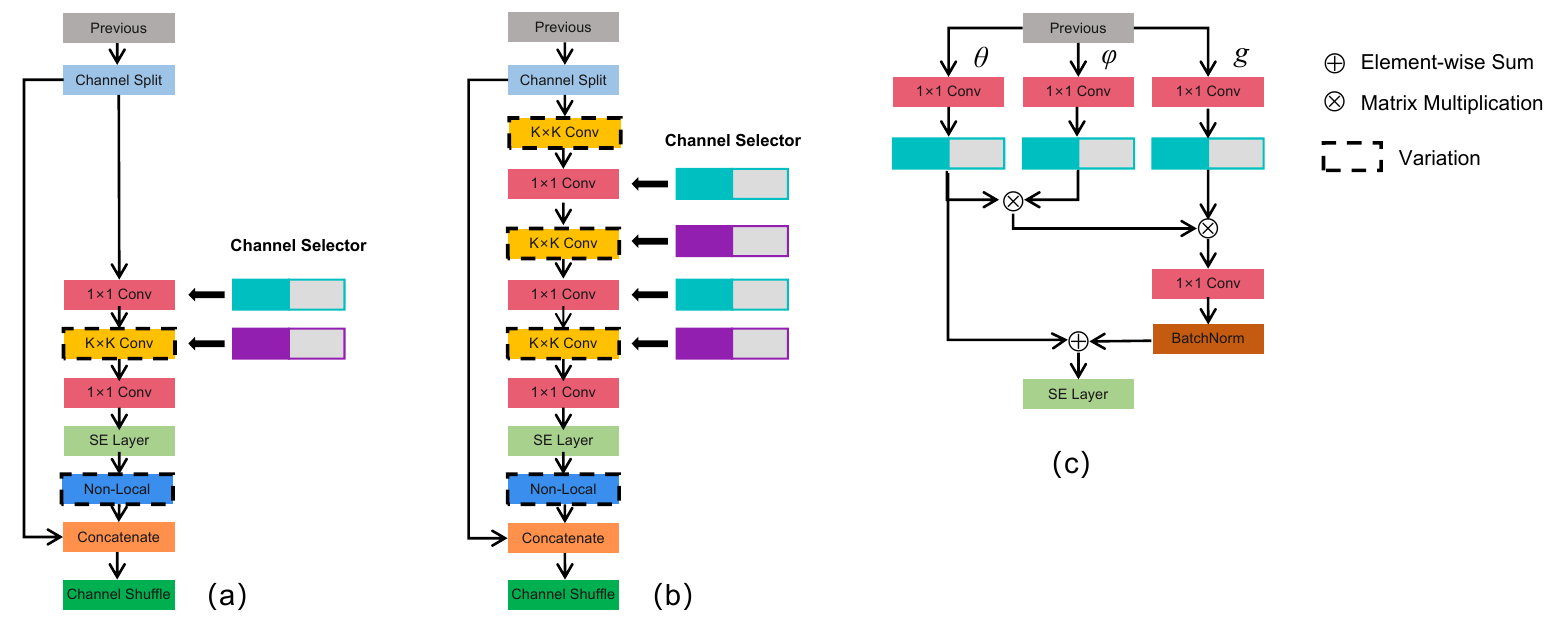}
\end{center}
\caption{(a), (b), (c) shows that how we add the channel selector into the tiny robust blocks and the pure robust blocks, respectively.  refer to the kernel size and it ranges among 3, 5, 7. The dashed line indicates the internal search space for the tiny robust blocks. So if we removed the non-local layer from the main branch of (a) and (b), (a) and (b) will denote as the pure tiny blocks. }
\label{fig:channel_selector_block}
\end{figure*}

The channel number plays an important role in the neural network's efficiency and computational cost. Apart from the adversarial accuracy and clean accuracy, we select the total number of parameters as the third objective. We only search the intermediate channel number of each block, including pure robust blocks, pure tiny blocks, and tiny robust blocks. Heuristically, the reduction of intermediate channel number will not give rise to the deterioration of adversarial accuracy and clean accuracy, which is well suited for the balance between the performance of neural networks and their model size. Specifically, Fig. \ref{fig:channel_selector_block} shows how we add a channel selector into the intermediate part of our candidate blocks. The channel selector ratio ranges from 0 to 2 and its interval is 0.2. The encoding method of choice channels is to assign an order from 0 to 10 for each channel selector ratio. 

\subsection{\textbf{Uniform Sampling}}
Our supernet sampling strategy is to sample choice blocks at first and then jointly sample choice blocks and choice channels once the warm-up training of the supernet is completed. Since we find out that our supernet is difficult to converge when we jointly sample block choices and channel choices in the beginning. We build up a parameter table in advance to speed up the sampling procedure. We will give more details about block sampling and channel sampling.

\subsubsection{\textbf{Block Sampling}}
Our pilot studies suggest that the tiny supernet size ranges from 1.5M to 4M. In the block sampling phase, we only search block choices for the supernet architecture and set a constraint that the number of the supernet parameters range from 1.823M to 2.375M. In all experiments, we train the supernet for 500 epochs in block sampling phases and sample a new architecture in every 20 epochs.

\subsubsection{\textbf{Block and Channel Jointly Sampling}}
The supernet jointly searches block and channel choices after the phase of block sampling. We refer to this phase as block and channel jointly sampling. We set another constraint that the number of the supernet parameters should range from 1.61M to 2.37M in this phase. In all experiments, we train the supernet for 500 epochs in block and channel jointly sampling phase and sample a new architecture in every 20 epochs.

\subsection{\textbf{Adversarial Training}}
We aim to explore the influence of network architecture on its robustness against adversarial attacks. So we focus on white box adversarial attacks bounded by $l_{\infty}$. As we all know, PGD \cite{madry:madry2017towards} adversarial training is computationally expensive and hard to converge. We follow \cite{zhang:zhang2019theoretically, zhang:zhang2019you} and adopt the TRADES-YOPO-m-n algorithm \cite{zhang:zhang2019you} to speed up our adversarial training. This work adopts the loss function in \cite{zhang:zhang2019theoretically}, which is described as below:
\begin{equation} \label{eq:trades}
    \min \mathbb{E}\left \{\mathcal{L}(f_{\theta}(x), y) + \max_{\left \|\eta \leq \epsilon  \right \|}\mathcal{L}(f_{\theta}(x), f_{\theta}(x + \eta))/\lambda \right \},
\end{equation}
\revised{where $\mathcal{L}\left ( .,. \right )$ denotes the cross-entropy loss function; $f_{\theta}(x)$ denotes the output vector of the neural network, which is parameterized by $\theta$. $y$ is the label-indicator vector; $\eta$ denotes the image noise (perturbation); $\lambda$ is a balancing hyperparameter. $f_{\theta}(x + \eta)$ denotes the output vector of the neural network where the perturbation $\eta$ are added into the input. The aim of this loss function is to reduce the gap between adversarial examples and non-adversarial examples when the model undertakes the classification task, i.e. making the classification boundary more smooth.} TRADES-YOPO-m-n borrows the idea from Pontryagin's Maximum Principle \cite{kopp1962pontryagin} to approximate the back-propogation. One assumption of TRADES-YOPO-m-n is that the adversarial perturbation is only coupled with the weights of the first layer. TRADES-YOPO-m-n performs \textit{n} times gradient descent to update the weights of the first layer and iteratively runs \textit{m} times for each data point. Zhang et al. \cite{zhang:zhang2019you} state that $m\times n$ should be a little larger than the number of attack iterations so that TRADES-YOPO-m-n could achieve a competitive result. In our experimental setting, the number of outer-loops \textit{m} is set to 5 and the number of inner loop \textit{n} is set to 3 if we desire to attack the models in 10 iterations by PGD \cite{madry:madry2017towards}. The training time for each epoch is 2.5 minutes running on a single GPU V100. The supernet adversarial training algorithm is presented in Algorithm \ref{supernet_training}.

\begin{algorithm}[!h]
\SetAlgoLined\small
\ForEach{ New Candidate Network $\in$ Block Sampling }{
\While{$epoch \leq 20$}{
Trades-YOPO-m-n\;\label{adv_training_block_sampling}

Update the corresponding weight of Supernet\;\label{update_weight_block_sampling}
}}\label{endupdate_block_sampling}
\BlankLine
\ForEach{ New Candidate Network $\in$ Block and Channel Jointly Sampling }{
\While{$epoch \leq 20$}{
Trades-YOPO-m-n\;\label{adv_training_block_channel_sampling}

Update the corresponding weight of Supernet\;\label{update_weight_block_channel_sampling}
}}\label{endupdate_block_channel_sampling}

\textbf{Return} Supernet
\caption{Supernet Training}\label{supernet_training}
\end{algorithm}

\begin{algorithm}[!t]
\SetAlgoLined 
\small
\SetKwInOut{Input}{Input}\SetKwInOut{Output}{Output}
\Input{population size $N$., crossover probability $p_{c}$, mutation probability $p_{m}$, number of generations $G$, supernet $S_{w}$,, clean accuracy predictor $S_{c}$, adversarial accuracy $S_{a}$, parameters calculator $S_{cp}$, \revised{$P^{0}$ the first generated population, $P_{g}$ the $g$th generated population, $F^{0}$ all non-dominated fronts of first generated population, $F_{g}$ all non-dominated fronts of $g$th generated population, $F_{g}^{0}$ the first non-dominated front of $g$th generated population, $F_{g}^{i}$ the $i$th non-dominated front of $g$th generated population. $Q_{0}$ the first offspring produced from the first population $P_{0}$, $Q_{g}$ the $g$th offspring produced from the $g$th population $P_{g}$}}
\Output{$F_{G}^{0}$ The first non-dominated front on $G$th generation}

$g \leftarrow 0$ \tcp{initialize a generation counter} \nllabel{gerneration_counter_initialize}

$P_{0} \leftarrow$ Initialize Population $N$ from Supernet $S_{w}$ \nllabel{population_initialize}

$f^{clean}_{p_{0}} \leftarrow S_{c}(P_{0})$ \tcp{compute clean accuracy for the first population} 

    $f^{adv}_{p_{0}} \leftarrow S_{a}(P_{0})$ \tcp{compute adversarial accuracy for the first population}

$f^{params}_{p_{0}} \leftarrow S_{cp}(P_{0})$ \tcp{calculate the size of models for the first population}

$F^{0} \leftarrow$ Fast Nondominated Sort$(P_{0})$ \nllabel{first_non_dominated_sort}

$F^{0} \leftarrow$ Crowding Distance Assignment($F^{0}$)

$P_{0} \leftarrow$ Binary Tournament Selection($P_{0}$) \tcp{choose parents through tournament selection for mating}

$Q_{0} \leftarrow$ Crossover($P_{0}, p_{c}$) \tcp{create offspring population by crossover between parents}

$Q_{0} \leftarrow$ Mutation($Q_{0}, p_{m}$) \tcp{induce randomness to offspring population through mutation}

$f^{clean}_{q_{0}} \leftarrow S_{c}(Q_{0})$ \tcp{compute clean accuracy for the $g+1$th offspring} 

$f^{adv}_{q_{0}} \leftarrow S_{a}(Q_{0})$ \tcp{compute adversarial accuracy for the $g+1$th offspring}

$f^{params}_{q_{0}} \leftarrow S_{cp}(Q_{0})$ \tcp{calculate the size of models for the $g+1$ offspring}

\While{$g < G$}{

$R_{g} \leftarrow P_{g} \:\cup\: Q_{g}$ \tcp{merge parent and offspring population}

$F_{g} \leftarrow$ Fast Nondominated Sort$(F_{g})$

$P_{g+1} \leftarrow$ 0 

$i \leftarrow 1$

\While{$P_{g+1} + F^{i}_{g} \leq N$}{
$F^{i}_{g} \leftarrow$ Crowding Distance Assignment($F^{i}_{g}$)\\
$P_{g+1} \leftarrow P_{g+1} \cup F^{i}_{g}$\\
$i = i + 1$
}

Sort($F^{i}_{g}$) \tcp{sort in descending order by the crowding distance operator}

$P_{g+1} \leftarrow P_{g+1} \cup F^{i}_{g}\left [ 1 : (N - P_{g+1}) \right ]$ \tcp{choose the first $(N - P_{g+1})$ elements of $F^{i}_{g}$}

$P_{g+1} \leftarrow$ Binary Tournament Selection($P_{g+1}$)

$Q_{g+1} \leftarrow$ Crossover($P_{g+1}, p_{c}$) 

$Q_{g+1} \leftarrow$ Mutation($Q_{g+1}, p_{m}$) 

\revised{$f^{clean}_{q_{g+1}} \leftarrow S_{c}(Q_{g+1})$ \tcp{compute clean accuracy for the first offspring}} 

\revised{$f^{adv}_{q_{g+1}} \leftarrow S_{a}(Q_{g+1})$ \tcp{compute adversarial accuracy for the first offspring}}

\revised{$f^{params}_{q_{g+1}} \leftarrow S_{cp}(Q_{g+1})$ \tcp{calculate the size of models for the first offspring}}
}

\textbf{Return} $F^{0}_{G}$
\caption{Multi-Objective Search}\label{nsga2}
\end{algorithm}

\subsection{\textbf{Multi-Objective Search}}
We use NSGA-II \cite{deb2002fast} as the multi-objective search algorithm. The multi-objective search algorithm is presented in Algorithm \ref{nsga2}. The first objective is the clean accuracy. It evaluates the performance of model without being attacked. The second objective is the adversarial accuracy. It evaluates the performance of model under white-box PGD attack bounded by $l_{\infty}$. The third objective is to evaluate the number of parameters of our searched
subnet. Moreover, the weight of each searched subnet clones from the corresponding part of the supernet. So there is no need to train any searched subnet in the whole search process. We are able to get the first objective value and second objective value after the inference of each searched subnet and we can quickly obtain the number of parameters for each searched subnet by checking our parameters table. The multi-objective search algorithm used in our framework is NSGA-II \cite{deb2002fast}. In the stage of initialization from Line 1 to Line 13 in Algorithm 2, we randomly initialize the parent population $P_{0}$ in the beginning where each individual (subnet) of population is evaluated with three fitness value: adversarial error, clean error and the number of parameters. The weight of each subnet clones from the corresponding part of supernet.

Then $P_{0}$ is sorted based on the non-dominated sorting. From Line 9 to Line 10 in Algorithm 2, we employ the tournament scheme and mutation operator to generate a offspring population $Q_{0}$. During the iteration of optimization (Lines 14-32), a combined population $R_{g} = P_{g} \cup Q_{g}$ is formed. Then $R_{g}$ will be sorted into fronts of individuals in an ascending order according to the non-dominated sorting, as shown in Line 16. Then new population $P_{g+1}$ is achieved from $R_{g}$ by selecting the elite solutions front by front according to their front number in an ascending order, as shown in Lines 19 - 23. The selection continues until $F_{g}^{i}$ where only a part of solutions are selected and they are selected according to the crowding distance values in a descending order. We will present the details of crossover and mutation operators in the next part.

\subsubsection{\textbf{Crossover}}\label{crossover_operator}
Before crossover, our multi-objective search \revised{adopts} the tournament selection for choosing two parents. In the experimental settings, the number of individuals that participate in tournament scheme is 10. Our crossover operator inherits and recombines the block or channel from the two parents to generate the new subnet. \revised{In order to solve the channel dimensional mismatch problems, our solution is to preallocate the weight matrix for the convolutional kernels ($ max\_output\_channel, max\_input\_channel, kernel\_size $). Our maximum channel dimension is two times as much as the original dimension. After crossover, the dimension of weight matrix for current batch is still unchanged. But we only keep the value of the weight matrix for current input and output channel $ [:c\_out, :c\_in, :] $. And the value of other channels in weight matrix are forced to be zero. In this case, we not only solve the channel dimension inconsistency problem, but also implement channel crossover and mutation conveniently. Moreover, we have to move non-local block out of the search space in stride-2 layer since the size of output feature map of non-local block cannot match the input size of the stride-1 layer}.

\subsubsection{\textbf{Mutation}}\label{mutation_operator}
The mutation operator is to re-assign each block or channel selector ratio of subnets from the search space. The mutation operator will be triggered if the randomized probability is larger than the pre-defined mutation probability. Our block encoding scheme is from 0 to 21. The block that contains non-local layer is from 12 to 21. \revised{Most of blocks can be arbitrarily mutated in each layer except for stride-2 layer. This is because the size of output feature map of non-local block cannot match the input size of the stride-1 layer. To enhance the diversity of population and prohibit creating completely different network architectures, we set the mutation probability to 0.1, which means the subnet has 10$\%$ opportunity to change the block or the channel number.}

\subsection{\textbf{Training From Scratch or Fine-Tuning}}
After finishing the multi-objective search, we obtain one set of non-dominated architectures. Generally speaking, there are two ways to deal with each searched subnet on the non-dominated front. One is to inherit the weight from the supernet and fine-tune, the other is to train it from scratch. We also use TRADES-YOPO-m-n for our adversarial training algorithm. We will examine the differences of these two approaches in the next section.

\section{Experimental Result and Analysis}
In this section, we introduce our experimental settings for the overall framework. In addition, we try to give a guideline for how to devise neural network architectures to defense adversarial attacks.  We perform extensive studies on CIFAR-10 \cite{krizhevsky2009learning} and SVHN \cite{netzer2011reading} to validate the effectiveness of our overall framework. On CIFAR-10, we do the zero-padding with the 4 pixels on each side and randomly crop back into the original size. Then we randomly flip the images horizontally and normalize them into $\left [0,1  \right ]$ for CIFAR and SVHN datasets. In order to better investigate the influence of the network architecture on robustness under adversarial attacks, we assume that the adversary has complete access to a neural network, including the architecture and all parameters. That is why we focus on white-box attacks on different architectures of neural networks.

\subsection{\textbf{Experimental Settings}}
\subsubsection{\textbf{Supernet}}
According to Algorithm \ref{supernet_training}, our supernet first enters into block sampling phase and the number of training epoch is set to 500. We provide 22 different blocks for the block sampling search space. We use the stochastic gradient descent method (SGD) as our optimizer. We use a batchsize of 512, a momentum of 0.9 and a weight decay of $5e-4$. The initial learning rate in block sampling is set to 0.1 and is lowered by 10 times at epoch 200, 400 and 450. After that, we jointly sample the blocks and channels of each layer in our supernet. We increase the number of epochs to 1000 with a batchsize of 512, a momentum of 0.9 and a weight decay of $5e-4$. We set our channel selector ratio to 1.8 and 2.0 before epoch 520 and add one more channel selector ratio 1.6 at epoch 540. The initial learning rate in block and channel jointly sampling is set to 0.1 and is lowered by 10 times at epoch 600, 700 and 800. The $\lambda$ of Equation \ref{eq:trades} is set to 1, which means that we try to balance the model performance on adversarial and non-adversarial examples.

\subsubsection{\textbf{NSGA-II}}
In our experiment settings, our total population size is 100. Then $P_{0}$ is sorted based on the non-dominated sorting and the size of $P_{0}$ is 50. From Line 9 to Line 10 in Algorithm 2, we employ the tournament scheme and mutation operator discussed in Sec. \ref{crossover_operator} and Sec. \ref{mutation_operator} to generate a offspring population $Q_{0}$ of size 50. The number of individuals which take part in tournament scheme is 10. During the iteration of optimization (Lines 14-32), a combined population $R_{t} = P_{t} \cup Q_{t}$ is formed and the size of $R_{t}$ is 100. Note that the population size of $P_{t+1}$ and $Q_{t+1}$ are both 50. The number of generation is set to 20. We use the hypervolume (HV) to indicate whether our search algorithm has been converged or not. Most of our experiments indicate that our multi-objective search algorithm has been converged at 18th generation. 

\subsubsection{\textbf{Training From Scratch}}
According to Fig. \ref{fig:framework}, we can obtain one set of non-dominated subnet architectures. We randomly initialize each subnet's weights and set the training epoch for every subnet to 100. But we find that most of subnets have converged at epoch 40. The initial learning rate of each subnet is 0.1 and is lowered by 10 times at epoch 20, 40 and 80. The optimizer we use here is SGD. We use a batchsize of 512, a momentum of 0.9 and a weight decay of $5e-3$. We evaluate our model on white-box $l_{\inf}$ bounded PGD attack with different number of epsilon and steps size. The epsilon for evaluation ranges from $2/255$ to $8/255$ and its interval is $2/255$. The number of PGD attack steps for evaluation ranges from 10 to 50. The hyperparameter of fine-tuning method is the same as mentioned above. Fine-tuning is to inherit the weight from the supernet for each subnet as initialization while training from scratch is to randomly initialize the weight of each subnet. 

\subsection{\textbf{Supernet Transferability}}
In this section, we aim to understand that if we use weaker PGD attack for the supernet adversarial training, whether it would largely deteriorate the adversarial performance of subnets. Specifically, we adjust the degree of PGD attack by changing the number of attack steps and epsilon size. To begin with, we build up a baseline for our best subnet in CIFAR-10 and SVHN dataset, which is presented in the first row of Table.  \ref{tab:supernet_transferability_number_of_attack_steps} and Table. \ref{tab:supernet_transferability_number_of_attack_steps_svhn}. The first column denotes how we train the supernet. For instance, \revised{subscript} of $\textbf{S}_{8/255}^{10}$ is the epsilon size of PGD attack for the supernet, which is set to $8/255$. And the \revised{superscript} is the number of attack steps, which is set to 10. The last column denotes that the best adversarial accuracy of subnet model under different degrees of attacks. For example, $\textbf{P}_{8/255}^{10}$ means the subnet is under PGD attack with epsilon size of $8/255$ and the number of attack steps of 10. Since we focus on the network architecture under adversarial attacks, the subnet presented in the following tables is the non-dominated architecture which achieves the best adversarial performance after training from scratch.

\subsubsection{\textbf{Number of Attack Steps}}

\begin{table*}[ht]
    \centering
    \setlength{\tabcolsep}{0.5mm}
    \caption{Supernet Transferability on Number of Attack Steps with the term of accuracy on CIFAR-10 (\%)}
    \begin{tabular}{c|c|c|cccccccccccc}
    Supernet Training& Subnet Model Size & Clean Acc & $\textbf{P}_{2/255}^{10}$ & $\textbf{P}_{4/255}^{10}$ & $\textbf{P}_{6/255}^{10}$ & $\textbf{P}_{8/255}^{10}$ & $\textbf{P}_{2/255}^{30}$ & $\textbf{P}_{4/255}^{30}$ & $\textbf{P}_{6/255}^{30}$ & $\textbf{P}_{8/255}^{30}$ & $\textbf{P}_{2/255}^{50}$ & $\textbf{P}_{4/255}^{50}$ & $\textbf{P}_{6/255}^{50}$ & $\textbf{P}_{8/255}^{50}$\\
    \hline
    $\textbf{S}_{8/255}^{10}$ & 1.7453M & 73.18 & 62.62 & 52.07 & 40.91 & 31.01 & 62.60 & 52.05 & 40.83 & 30.68 & 62.60 & 52.04 & 40.77 & 30.65 \\
    \hline
    $\textbf{S}_{8/255}^{1}$ & 1.7574M & 73.19 & 62.53 & 51.23 &39.83 & 29.35 & 62.55  & 51.23 & 39.77 & 29.01 & 62.52 & 51.20 & 39.77 & 28.92 \\
    \hline
    $\textbf{S}_{8/255}^{2}$ & \textbf{1.6822M} & \textbf{76.54} & \textbf{66.64} & \textbf{55.11} & \textbf{42.90}  & \textbf{31.83} & \textbf{66.65}  & \textbf{55.08}  & \textbf{42.73}  & \textbf{31.27} & \textbf{66.61} & \textbf{55.10} & \textbf{42.71} & \textbf{31.26} \\
    \hline
    \end{tabular}\label{tab:supernet_transferability_number_of_attack_steps}
\end{table*}

\begin{table*}[ht]
    \centering
    \setlength{\tabcolsep}{0.5mm}
    \caption{Supernet Transferability on Number of Attack Steps with the term of accuracy on SVHN (\%)}
    \begin{tabular}{c|c|c|cccccccccccc}
    Supernet Training& Subnet Model Size & Clean Acc & $\textbf{P}_{2/255}^{10}$ & $\textbf{P}_{4/255}^{10}$ & $\textbf{P}_{6/255}^{10}$ & $\textbf{P}_{8/255}^{10}$ & $\textbf{P}_{2/255}^{30}$ & $\textbf{P}_{4/255}^{30}$ & $\textbf{P}_{6/255}^{30}$ & $\textbf{P}_{8/255}^{30}$ & $\textbf{P}_{2/255}^{50}$ & $\textbf{P}_{4/255}^{50}$ & $\textbf{P}_{6/255}^{50}$ & $\textbf{P}_{8/255}^{50}$\\
    \hline
    $\textbf{S}_{8/255}^{10}$ & 1.7657M & 87.26 & 72.65 & 68.17 & 60.82 & 50.91 & 72.27 & 68.09 & 60.81 & 50.64 & 72.21 & 68.04 & 50.35 & 50.24 \\
    \hline
    $\textbf{S}_{8/255}^{1}$ & 1.7428M & 88.34 & 71.53 & 66.45 & 59.83 & 49.53 & 71.52 & 66.31 & 58.80 & 48.56 & 71.51 & 66.21 & 58.70 & 49.51 \\
    \hline
    $\textbf{S}_{8/255}^{5}$ & \textbf{1.6836M} & \textbf{90.54} & \textbf{76.28} & \textbf{70.16} & \textbf{62.73}  & \textbf{54.70} &  \textbf{76.15}  & \textbf{70.15}  & \textbf{62.11} & \textbf{54.69} & \textbf{76.10} & \textbf{70.15} & \textbf{62.09} & \textbf{54.68} \\
    \hline
    \end{tabular}\label{tab:supernet_transferability_number_of_attack_steps_svhn}
\end{table*}

Table. \ref{tab:supernet_transferability_number_of_attack_steps} indicates that the subnet performs better even if the supernet is under fewer number of attack steps during the adversarial training. In comparison with $S_{8/255}^{10}$, $S_{8/255}^{2}$ helps to increase the adversarial accuracy and clean accuracy by 6.4$\%$ and 4.5$\%$, respectively, and reduce its subnet model size by 3.6$\%$. In addition, the gap between $S_{8/255}^{1}$ and our baseline $S_{8/255}^{10}$ is very tiny for different objectives. So we surmise that the supernet will have strong transferability even we reduce the number of attack steps for its own adversarial training. Moreover, we can easily observe that the adversarial accuracy of subnets is strongly affected by the epsilon size but not the number of attack steps. For example, the difference of the adversarial accuracy $P_{2/255}^{10}$, $P_{2/255}^{30}$ and  $P_{2/255}$ is very small regardless how the supernet is trained. It meets the same conclusion in Table. \ref{tab:supernet_transferability_number_of_attack_steps_svhn}. Therefore, we think that this observation not only helps us to save much more training time by reducing the number of attack steps but also make it possible for the subnets to obtain stronger adversarial defensive ability.

\subsubsection{\textbf{Epsilon Size}}
Table. \ref{tab:supernet_transferability_epsilon_size} indicates that the supernet is able to improve its subnets' representation abilities if it is not overloaded with the epsilon size. Firstly, the subnet of $S_{2/255}^{10}$ achieves the highest clean accuracy up to 81.95$\%$, but the adversarial accuracy of its subnet under $P_{8/255}^{10}$, $P_{8/255}^{30}$, $P_{8/255}^{50}$ attacks is unable to surpass 9$\%$. Our assumption is that the subnet has not fully developed its resilience ability since its supernet is incapable of learning by generating adversarial examples with a larger epsilon size. The assumption has been verified that the subnet of $S_{4/255}^{10}$ and $S_{6/255}^{10}$ hugely increase the adversarial accuracy when it is under $P_{8/255}^{10}$ attack.  Another assumption is that if the epsilon size exceeds the supernet's workload, it will reduce both the clean accuracy and adversarial accuracy of subnets. We can easily observe that the subnet of $S_{6/255}^{10}$ obtains the best adversarial performance while $S_{8/255}^{10}$ largely weakens its subnet's performance regardless in clean accurarcy, or under $P_{8/255}^{10}$, $P_{6/255}^{10}$, $P_{4/255}^{10}$ attacks. However, when the epsilon size exceeds $6/255$ in our case, the subnet performance begins to decline gradually. It meets the same conclusion in Table. \ref{tab:supernet_transferability_epsilon_size_svhn}.

\begin{table*}[ht]
    \centering
    \setlength{\tabcolsep}{0.5mm}
    \caption{Supernet Transferability on Epsilon Size with the term of accuracy on CIFAR-10 (\%)}
    \begin{tabular}{c|c|c|cccccccccccc}
    Supernet Training& Subnet Model Size & Clean Acc & $\textbf{P}_{2/255}^{10}$ & $\textbf{P}_{4/255}^{10}$ & $\textbf{P}_{6/255}^{10}$ & $\textbf{P}_{8/255}^{10}$ & $\textbf{P}_{2/255}^{30}$ & $\textbf{P}_{4/255}^{30}$ & $\textbf{P}_{6/255}^{30}$ & $\textbf{P}_{8/255}^{30}$ & $\textbf{P}_{2/255}^{50}$ & $\textbf{P}_{4/255}^{50}$ & $\textbf{P}_{6/255}^{50}$ & $\textbf{P}_{8/255}^{50}$\\
    \hline
    $\textbf{S}_{8/255}^{10}$ & 1.7453M & 73.18 & 62.62 & 52.07 & 40.91 & \textbf{31.01} & 62.60 & 52.05 & 40.83 & \textbf{30.68} & 62.60 & 52.04 & 40.77 & \textbf{30.65} \\
    \hline
    $\textbf{S}_{6/255}^{10}$ & 1.6840M & 78.63 & \textbf{67.79} & \textbf{54.85} & \textbf{41.584} & 29.03 & \textbf{67.76} & \textbf{54.80} & \textbf{41.42} & 28.19 & \textbf{67.78} & \textbf{54.80} & \textbf{41.32}  & 28.09 \\
    \hline
    $\textbf{S}_{4/255}^{10}$ & 1.6775M & 75.82 & 62.63 & 46.87 & 31.87 & 20.27 & 62.65 & 46.85 & 31.65 & 19.62 & 62.62 & 46.86 & 31.65 & 19.55 \\
    \hline
    $\textbf{S}_{2/255}^{10}$ & \textbf{1.6632M} & \textbf{81.95} & 62.14 & 38.52 & 20.29 & 8.823 & 62.15 & 38.21 & 19.60 & 7.675 & 62.13 & 38.18 & 19.54 & 7.606 \\
    \hline
    \end{tabular}\label{tab:supernet_transferability_epsilon_size}
\end{table*}

\begin{table*}[ht]
    \centering
    \setlength{\tabcolsep}{0.5mm}
    \caption{Supernet Transferability on Epsilon Size with the term of accuracy on SVHN (\%)}
    \begin{tabular}{c|c|c|cccccccccccc}
    Supernet Training& Subnet Model Size & Clean Acc & $\textbf{P}_{2/255}^{10}$ & $\textbf{P}_{4/255}^{10}$ & $\textbf{P}_{6/255}^{10}$ & $\textbf{P}_{8/255}^{10}$ & $\textbf{P}_{2/255}^{30}$ & $\textbf{P}_{4/255}^{30}$ & $\textbf{P}_{6/255}^{30}$ & $\textbf{P}_{8/255}^{30}$ & $\textbf{P}_{2/255}^{50}$ & $\textbf{P}_{4/255}^{50}$ & $\textbf{P}_{6/255}^{50}$ & $\textbf{P}_{8/255}^{50}$\\
    \hline
    $\textbf{S}_{8/255}^{10}$ & 1.7657M & 87.26 & 72.65 & 68.17 & 60.82 & \textbf{50.91} & 72.27 & 68.09 & 60.81 & \textbf{50.64} & 72.21 & 68.04 & 60.35 & \textbf{50.24} \\
    \hline
    $\textbf{S}_{6/255}^{10}$ & 1.6947M & 88.45 & \textbf{72.37} & \textbf{69.46} & \textbf{61.47} & 49.87 & \textbf{72.26} & \textbf{69.32} & \textbf{61.42} & 49.82 & \textbf{72.15} & \textbf{69.27} & \textbf{61.32}  & 49.85 \\
    \hline
    $\textbf{S}_{4/255}^{10}$ & 1.6873M & 85.82 & 62.37 & 58.47 & 52.43 & 42.70 & 62.35 & 58.37 & 51.49 & 41.50 & 62.30 & 58.21 & 51.37 & 40.61 \\
    \hline
    $\textbf{S}_{2/255}^{10}$ & \textbf{1.6712M} & \textbf{91.14} & 60.14 & 50.66 & 39.38 & 21.50 & 59.88 & 49.53 & 38.27 & 20.17 & 58.96 & 48.18 & 37.12 & 20.54 \\
    \hline
    \end{tabular}\label{tab:supernet_transferability_epsilon_size_svhn}
\end{table*}

\begin{table*}[ht]
    \centering
    \setlength{\tabcolsep}{0.5mm}
    \caption{Supernet Transferability on Model Size with the term of accuracy on CIFAR-10 (\%)}
    \begin{tabular}{c|c|c|cccccccccccc}
    Supernet Training& Subnet Model Size & Clean Acc & $\textbf{P}_{2/255}^{10}$ & $\textbf{P}_{4/255}^{10}$ & $\textbf{P}_{6/255}^{10}$ & $\textbf{P}_{8/255}^{10}$ & $\textbf{P}_{2/255}^{30}$ & $\textbf{P}_{4/255}^{30}$ & $\textbf{P}_{6/255}^{30}$ & $\textbf{P}_{8/255}^{30}$ & $\textbf{P}_{2/255}^{50}$ & $\textbf{P}_{4/255}^{50}$ & $\textbf{P}_{6/255}^{50}$ & $\textbf{P}_{8/255}^{50}$\\
    \hline
    $\textbf{S}_{8/255}^{10}$ & 1.7453M & 73.18 & 62.62 & 52.07 & 40.91 & 31.01 & 62.60 & 52.05 & 40.83 & 30.68 & 62.60 & 52.04 & 40.77 & 30.65 \\
    \hline
    $\textbf{S}_{6/255}^{10}$ & 1.6840M & \textbf{78.63} & \textbf{67.79} & 54.85 & 41.584 & 29.03 & \textbf{67.76} & 54.80 & 41.42 & 28.19 & \textbf{67.78} & 54.80 & 41.32 & 28.09 \\
    \hline
    $\textbf{S}_{8/255}^{2}$ & \textbf{1.6822M} & 76.54 & 66.64 & \textbf{55.11} & \textbf{42.90}  & \textbf{31.83} & 66.65  & \textbf{55.08}  & \textbf{42.73}  & \textbf{31.27} & 66.61 & \textbf{55.10} & \textbf{42.71} & \textbf{31.26} \\
    \hline
    \end{tabular}\label{tab:model_size_effect}
\end{table*}

\subsection{\textbf{The Model Size}}
Table. \ref{tab:model_size_effect} shows that the adversarial performance of model may not be closely correlated with the model size, which is inconsistent with the conclusion in \cite{madry:madry2017towards}. When comparing $S_{8/255}^{2}$ with our baseline model $S_{8/255}^{10}$, the subnet model size drops by 3.751$\%$ but the clean accuracy and adversarial accuracy in $P_{8/255}^{10}$, $P_{6/255}^{10}$, $P_{4/255}^{10}$, $P_{2/255}^{10}$ separately increase by 4.591\%, 2.642\%, 4.864\%, 5.838\%, 6.420\%. When comparing with the subnet of $S_{6/255}^{10}$ and $S_{8/255}^{10}$, we find that the best subnet size drops by 3.640\%, but the clean accuracy and the adversarial accuracy in $P_{6/255}^{10}$, $P_{4/255}^{10}$, $P_{2/255}^{10}$ separately increase by 7.447\%, 1.63\%, 5.338\%, 8.256\%. So we think that the performance of subnet has strong correlation with the training mode of its supernet but not the mode size of itself.

\subsection{\textbf{Training From Scratch or Fine Tuning}}

\begin{figure*}[htb]
\centering
\subfigure[$S_{8/255}^{10}$]{
    \label{fig:step_10}
    \includegraphics[width=0.3\textwidth]{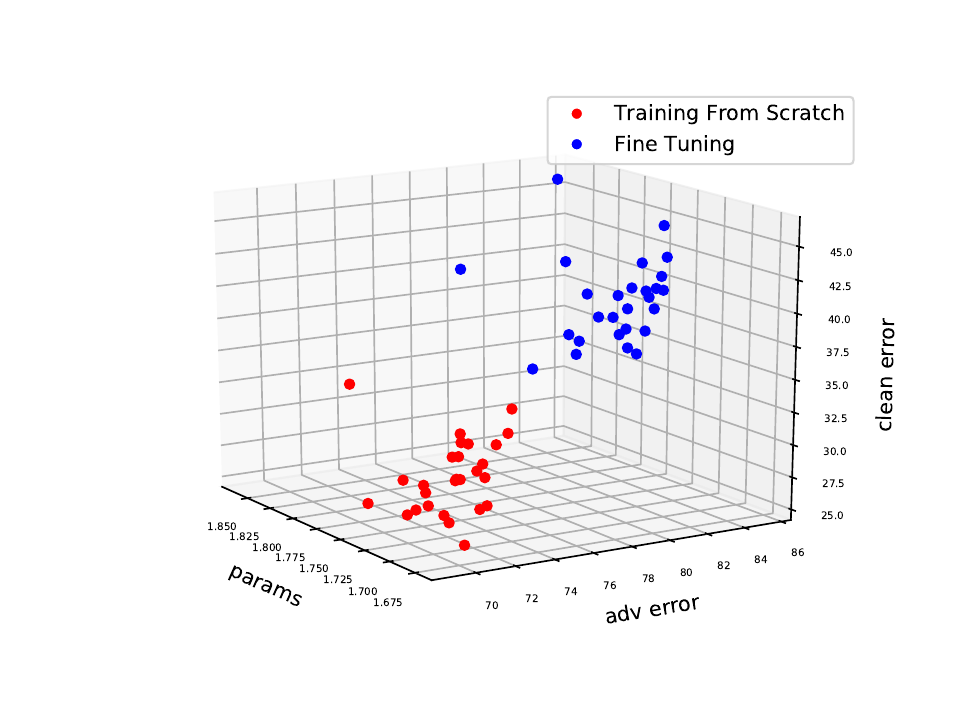}}
\subfigure[$S_{8/255}^{2}$]{
    \label{fig:step_2}
    \includegraphics[width=0.3\textwidth]{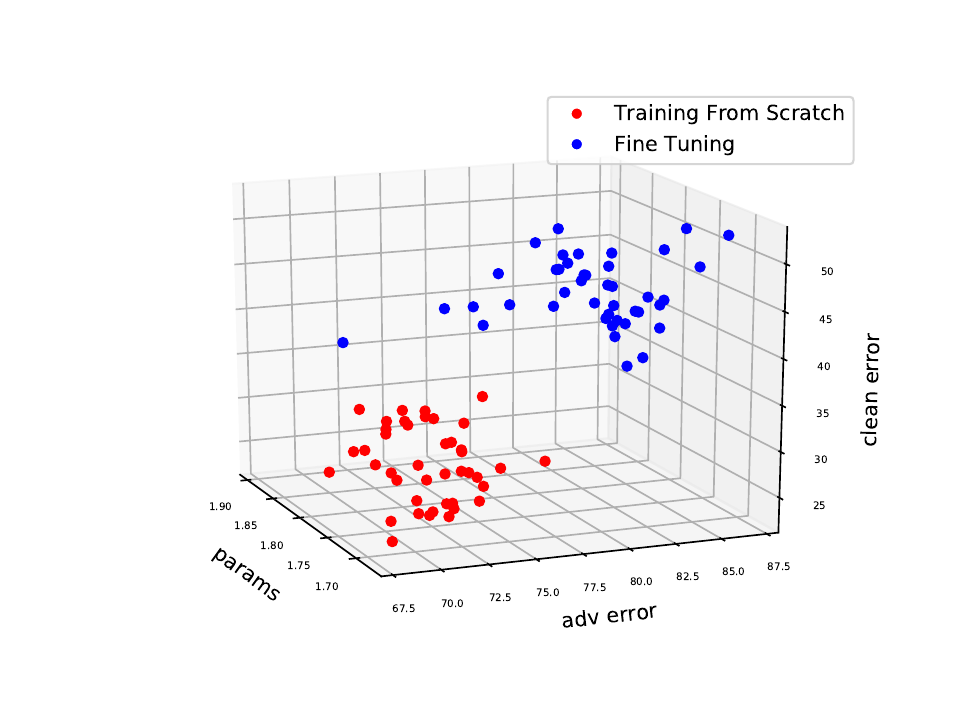}}
\subfigure[$S_{4/255}^{10}$]{
    \label{fig:ep_4_step_10}
    \includegraphics[width=0.3\textwidth]{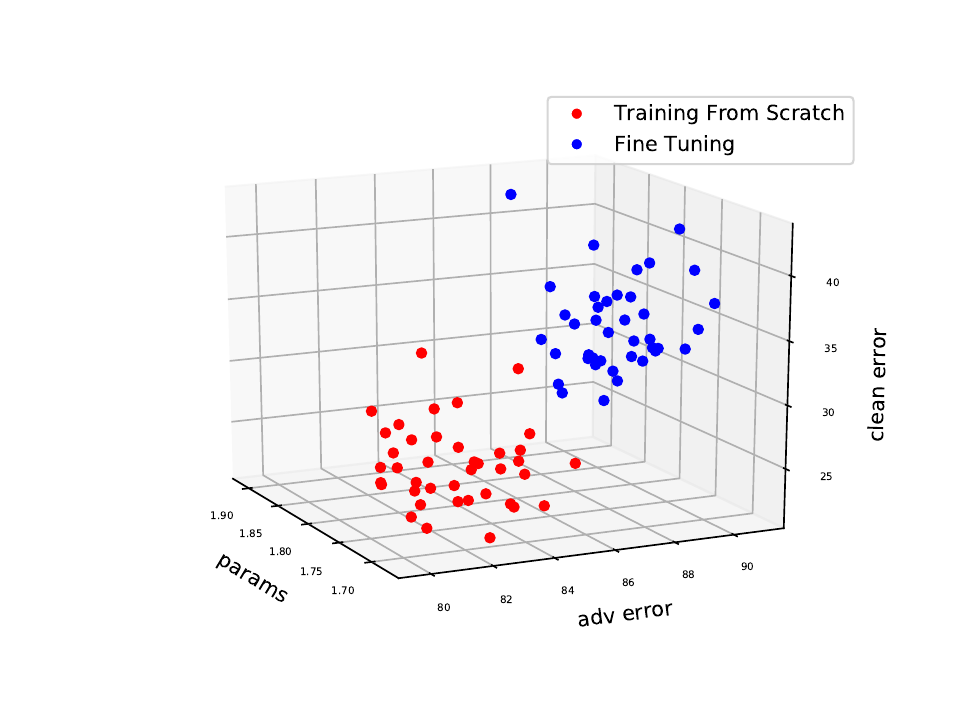}}
\caption{Difference between training from scratch and fine-tuning. The red points represent the solutions on the non-dominated front obtained by NSGA-II, where the weights for subnets are randomly initialized. By contrast, the blue points are different solutions achieved by the same way but the weights of the subnets are inherited from supernet.}
\label{fig:difference_train_from_scratch_finetune}
\end{figure*}

We can easily observe there does exists the huge gap between training from scratch and fine-tuning from Fig. \ref{fig:difference_train_from_scratch_finetune}. The three axes for each graph represent the number of parameters, the adversarial error and clean error for each subnet, respectively. The subscript for each graph in Fig. \ref{fig:difference_train_from_scratch_finetune} denotes the supernet training mode. The red points represent the solutions on the non-dominated front obtained by NSGA-II, where the weights for subnets are randomly initialized. By contrast, the blue points are different solutions achieved by the same way but the weights of the subnets are inherited from supernet. We can clearly observe that no matter how we train the supernet, the subnets which \revised{adopt} training from scratch as initialization mode perform better on adversarial examples and non-adversarial examples. We hypothesize that the role of the supernet in our NAS framework is to find the best architecture for the subnet but not to deliver the best weight to the subnet. Fine-tuning is not always beneficial to the training. The reason may be that the weight of each newly sampled subnet is not good enough as there are only 20 epochs for the training. However, we can find the best subnets among them by means of NSGA-II during the optimization in terms of the objectives.

\subsection{\textbf{Subnets Analysis}}
This section analyzes the top ten subnets architecture in terms of the clean accuracy and adversarial accuracy and their counterpart supernet training method on CIFAR-10 and SVHN datasets. Our aim is to gain insights from our top best results and reveal the rule for how to design a more robust tiny neural network.
\subsubsection{\textbf{Adversarial Error, Clean Error and the Size of Neural Network}}
\begin{figure*}[htb]
\centering
\subfigure[$2S_{8/255}^{1}$]{
    \label{fig:2c_step_1_cifar10_nondominated_front}
    \includegraphics[width=0.3\textwidth]{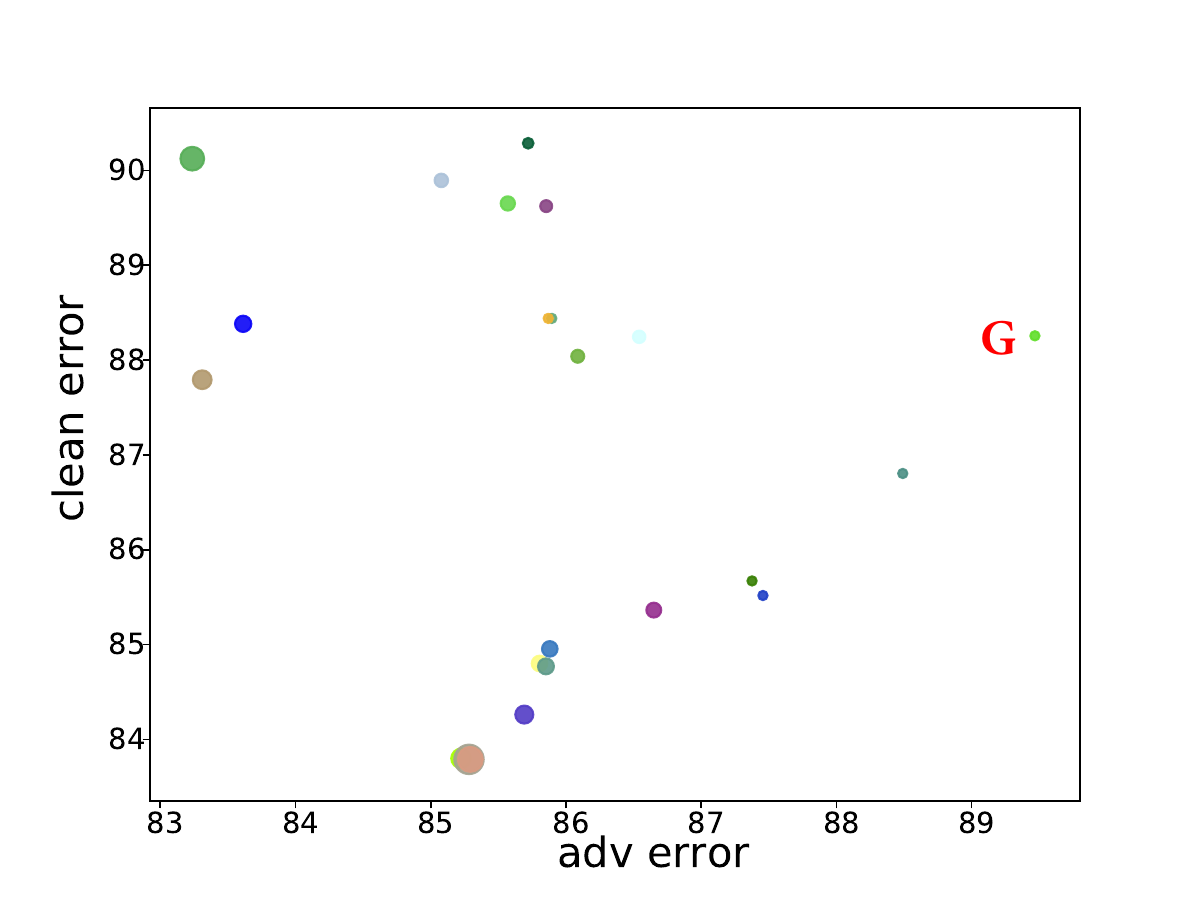}}
        \hspace{-5mm}
\subfigure[$S_{6/255}^{10}$]{
    \label{fig:ep_3_step_10_cifar10_nondominated_front}
    \includegraphics[width=0.3\textwidth]{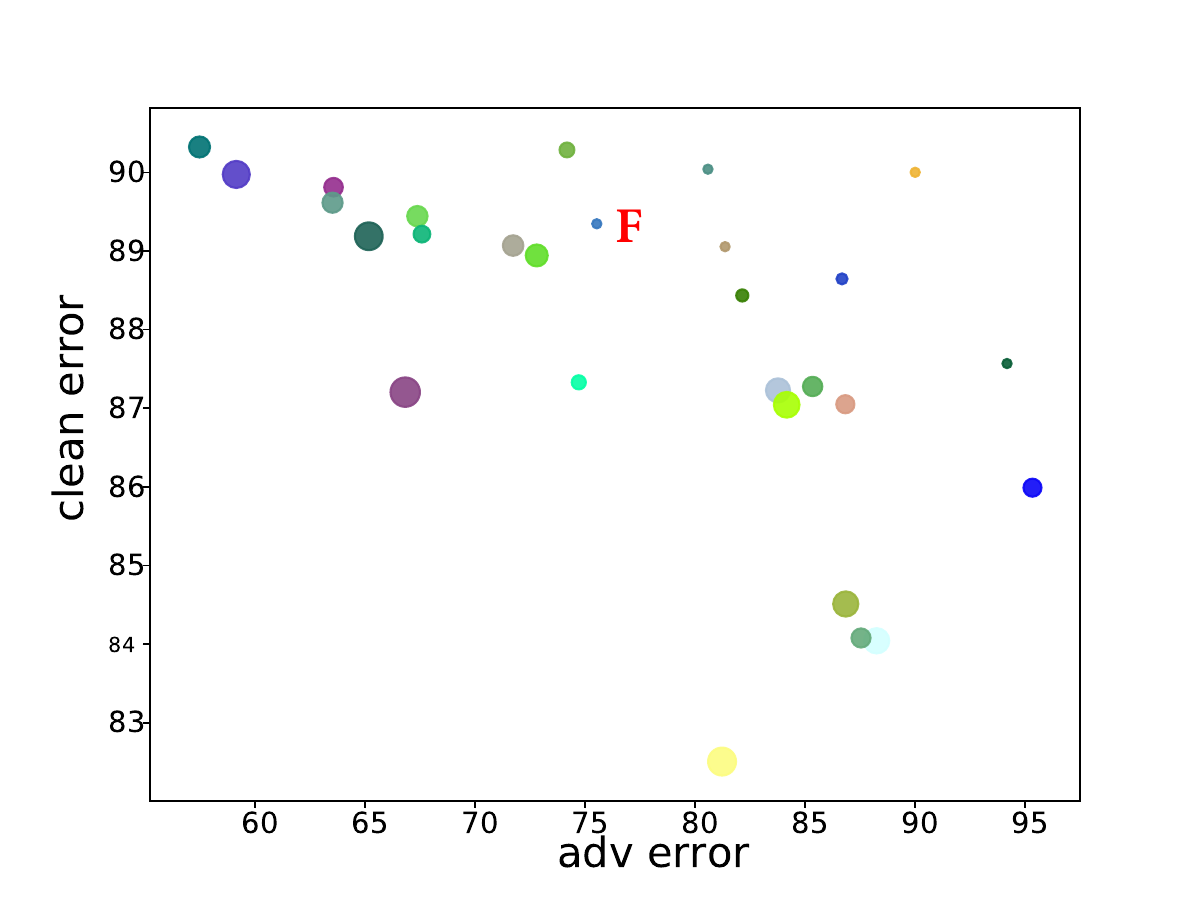}}
        \hspace{-5mm}
\subfigure[$S_{8/255}^{2}$]{
    \label{fig:step_2_cifar10_nondominated_front}
    \includegraphics[width=0.3\textwidth]{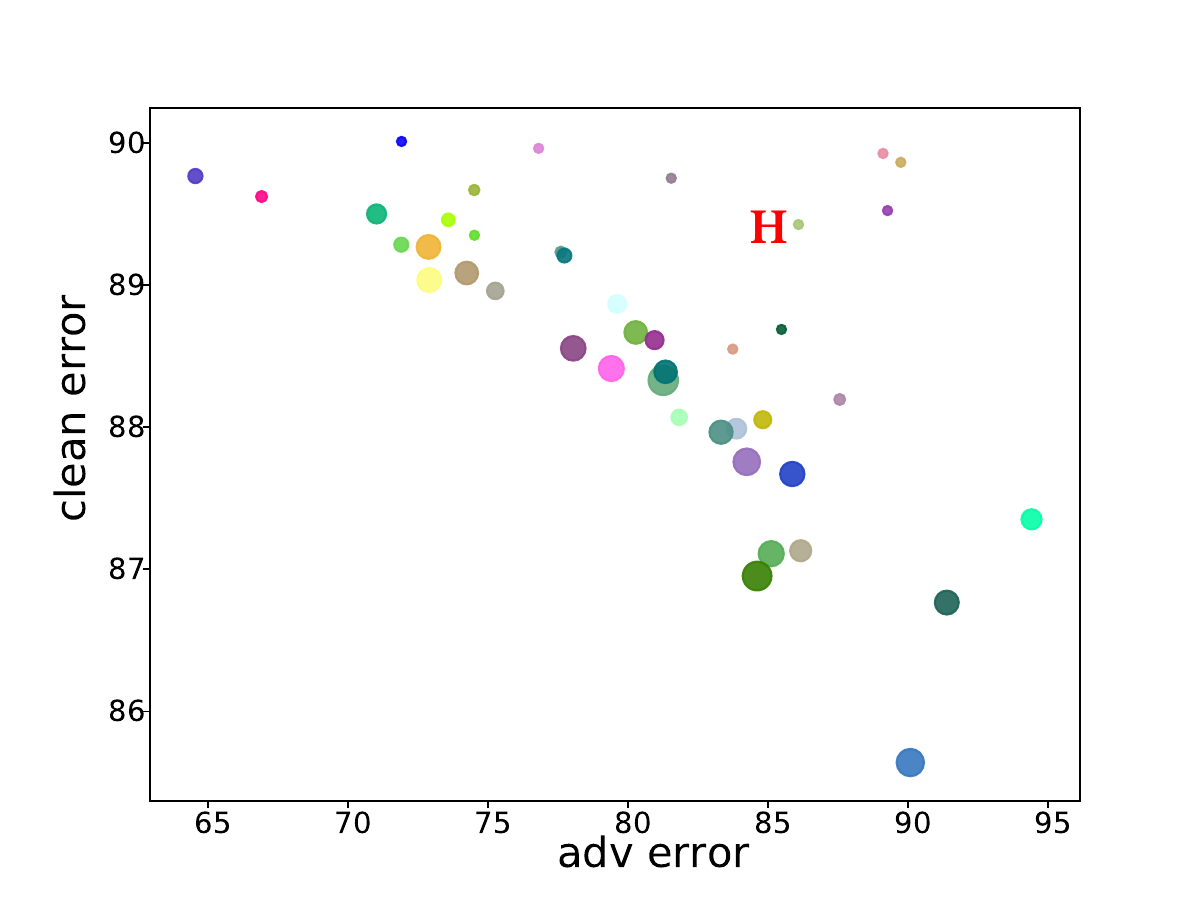}}
\caption{The first nondominated front before training from scratch among adversarial error, clean error and the number of parameters of subnets on CIFAR-10. The size of circle indicates the amount of parameter of subnets. The vertical axis represents the clean error of subnets while the horizontal axis represents the adversarial error of subnets.}
\label{fig:the_first_nondominated_front_cifar10}
\end{figure*}

\begin{figure*}[htb]
\centering
\subfigure[$2S_{8/255}^{1}$]{
    \label{fig:2c_step_1_cifar10_training_from_scratch}
    \includegraphics[width=0.3\textwidth]{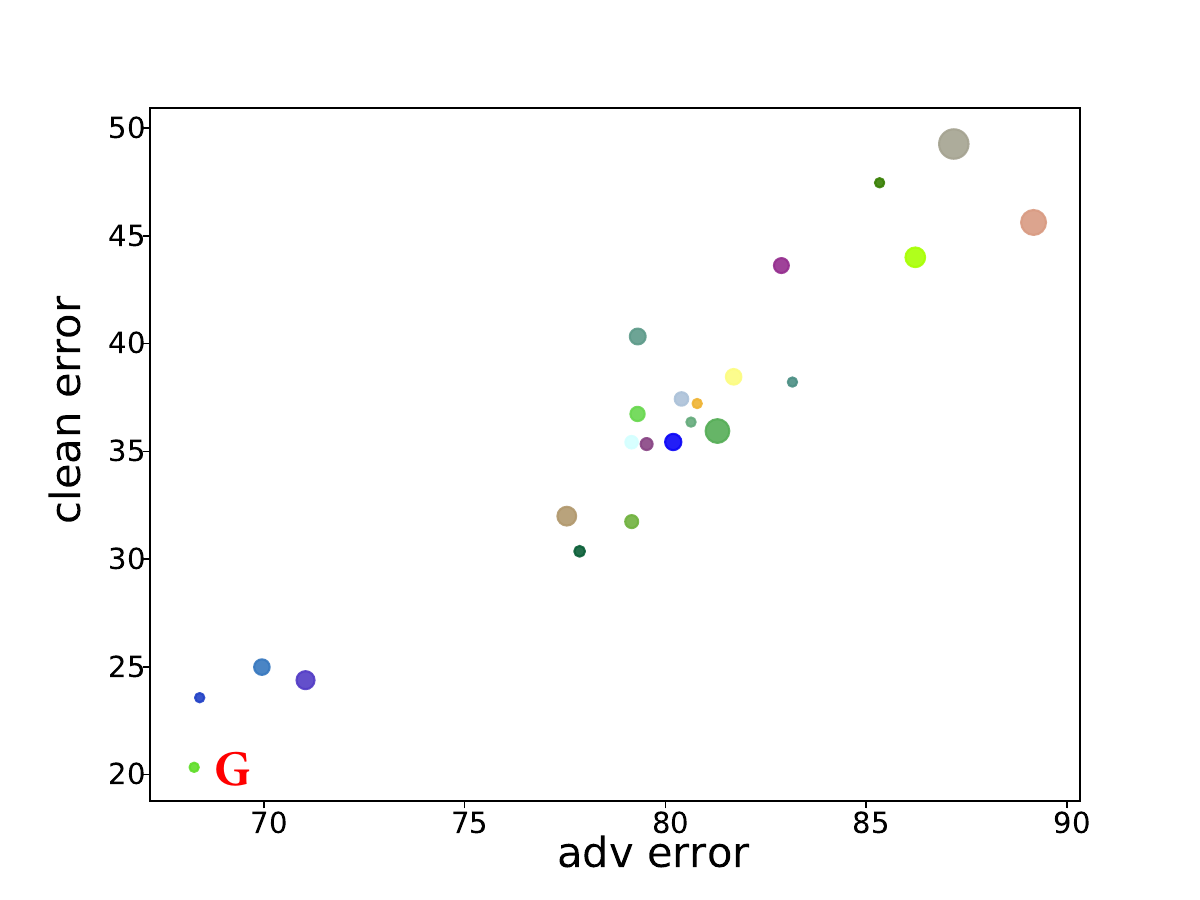}}
        \hspace{-5mm}
\subfigure[$S_{6/255}^{10}$]{
    \label{fig:ep_3_step_10_cifar10_training_from_scratch}
    \includegraphics[width=0.3\textwidth]{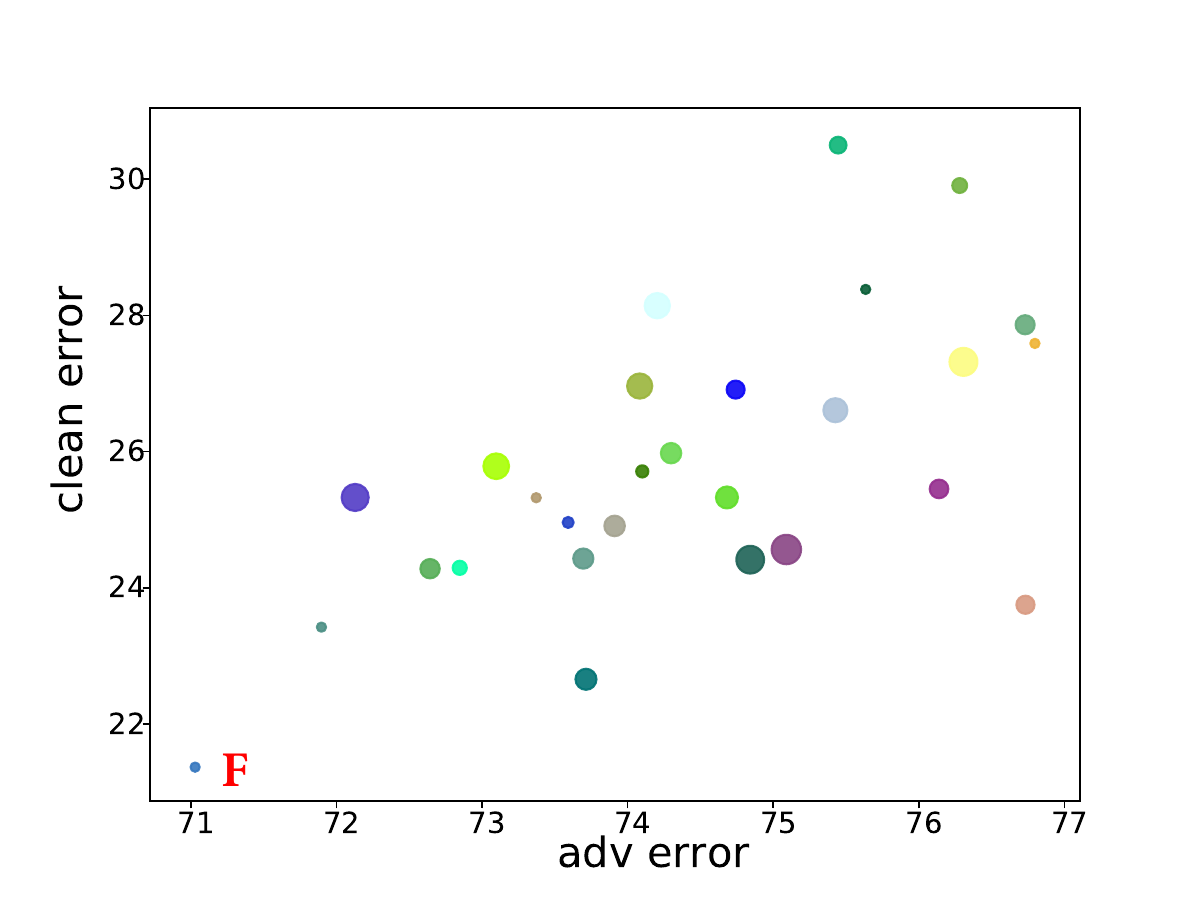}}
        \hspace{-5mm}
\subfigure[$S_{8/255}^{2}$]{
    \label{fig:step_2_cifar10_training_from_scratch}
    \includegraphics[width=0.3\textwidth]{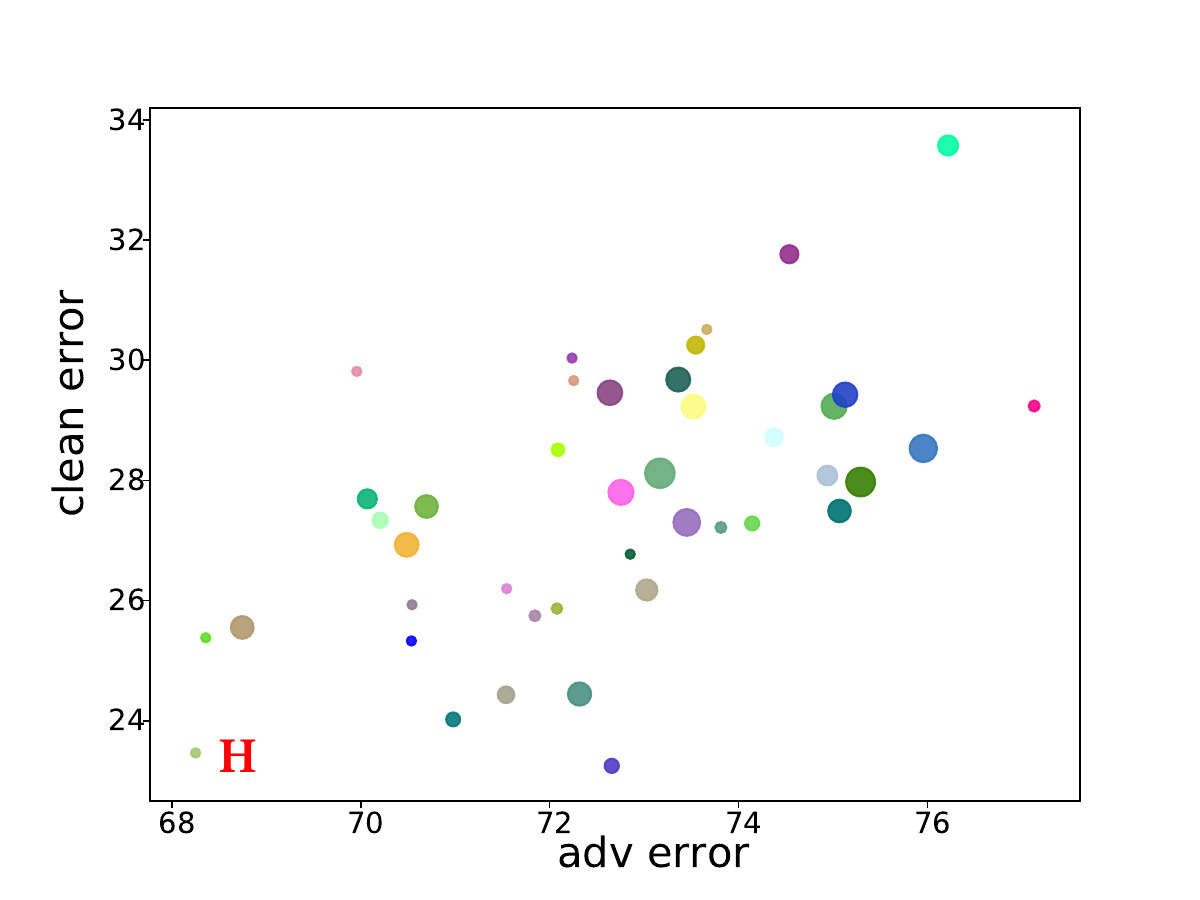}}
\caption{The first nondominated front after training from scratch on CIFAR-10. The size of circle indicates the amount of parameter of subnets. The vertical axis represents the clean error of subnets while the horizontal axis represents the adversarial error of subnets. The same color of circles in Fig.5 and Fig.6 means they hold the same network architecture.}
\label{fig:adversarial_clean_train_from_scratch_cifar10}
\end{figure*}

Fig. \ref{fig:the_first_nondominated_front_cifar10} shows that the nondominated front obtained by NSGA-II on CIFAR-10 dataset and their counterpart supernet come from $2S_{8/255}^{1}$, $S_{6/255}^{10}$ and $S_{8/255}^{2}$, respectively. We use circles to represent the subnets and the size of circle indicates its size (number of parameters). In Figs.6 (b) and (c), it can be easily observed that there does exist the trade-off relationship between adversarial error, clean error and the size of neural network. Fig. \ref{fig:adversarial_clean_train_from_scratch_cifar10} shows that the order of non-dominated subnets has greatly changed after training from scratch. In order to clearly illustrate how the order of non-dominated subnets changes after training from scratch, each subnet (circle) is denoted by different color. The same color circles in Fig. \ref{fig:the_first_nondominated_front_cifar10} and Fig. \ref{fig:adversarial_clean_train_from_scratch_cifar10} indicate that they own the same network architecture. We get an important observation from Fig. \ref{fig:the_first_nondominated_front_cifar10} and Fig. \ref{fig:adversarial_clean_train_from_scratch_cifar10} that most of tiny neural networks (tiny circles) achieve a significant reduction on both adversarial error and clean error after training from scratch. For instance, the G point in Fig. \ref{fig:2c_step_1_cifar10_training_from_scratch} which owns the lowest clean error and lowest adversarial error has larger adversarial error and clean error before training from scratch. It also meets the same observation for the F point when we compare with Fig. \ref{fig:ep_3_step_10_cifar10_nondominated_front} and Fig. \ref{fig:ep_3_step_10_cifar10_training_from_scratch}. Hence, we conclude that our pipeline can effectively increase adversarial accuracy and clean accuracy of the tiny neural networks.

\begin{table*}
    \centering 
    \caption{Block encoding scheme and layer statistics on CIFAR-10. Shuffle+SE: S, Shuffle-Xception+SE: SX, Non-Local-EmbeddedGaussian+SE: NE, Non-Local-Gaussian+SE: NG, BatchNorm: BN; K: kernel size.}
    \begin{tabular}{c|c|c|c|c|c|c|c|c|c|c|c|c|c|c|c|c|c|c|c|c|c|c}
    \hline \hline \small
     & \multicolumn{22}{c}{\textbf{Architecture}} \\ \hline
        \textbf{Block} & 0 & 1 & 2 & 3 & 4 & 5 & 6 & 7 & 8 & 9 & 10 & 11 & 12 & 13 & 14 & 15 & 16 & 17 & 18 & 19 & 20 & 21 \\ \hline 
        S & \checkmark & \checkmark & \checkmark &  &  &  & \checkmark & \checkmark & \checkmark &  &  &  & \checkmark & \checkmark & \checkmark &  &  &  &  &  &  &  \\ \hline
        SX &  &  &  & \checkmark & \checkmark & \checkmark &  &  &  & \checkmark & \checkmark & \checkmark &  &  &  & \checkmark & \checkmark & \checkmark &  &  &  &  \\ \hline
        NE &  &  &  &  &  &  & \checkmark & \checkmark & \checkmark & \checkmark & \checkmark & \checkmark &  &  &  &  &  &  & \checkmark & \checkmark &  &  \\ \hline
        NG &  &  &  &  &  &  &  &  &  &  &  &  & \checkmark & \checkmark & \checkmark & \checkmark & \checkmark & \checkmark &  &  & \checkmark & \checkmark \\ \hline
        BN &  &  &  &  &  &  &  &  &  &  &  &  &  &  &  &  &  &  & \checkmark &  & \checkmark &  \\ \hline
        K=3 & \checkmark &  &  & \checkmark &  &  & \checkmark &  &  & \checkmark &  &  & \checkmark &  &  & \checkmark &  &  &  &  &  &  \\ \hline
        K=5 &  & \checkmark &  &  & \checkmark &  &  & \checkmark &  &  & \checkmark &  &  & \checkmark &  &  & \checkmark &  &  &  &  &  \\ \hline
        K=7 &  &  & \checkmark &  &  & \checkmark &  &  & \checkmark &  &  & \checkmark &  &  & \checkmark &  &  & \checkmark &  &  &  &  \\ \hline 
    \end{tabular}
    \label{tab:block_encoding_statistics_cifar10}
\end{table*}

\begin{figure*}[htb]
    \centering
    \includegraphics[width=0.9\textwidth]{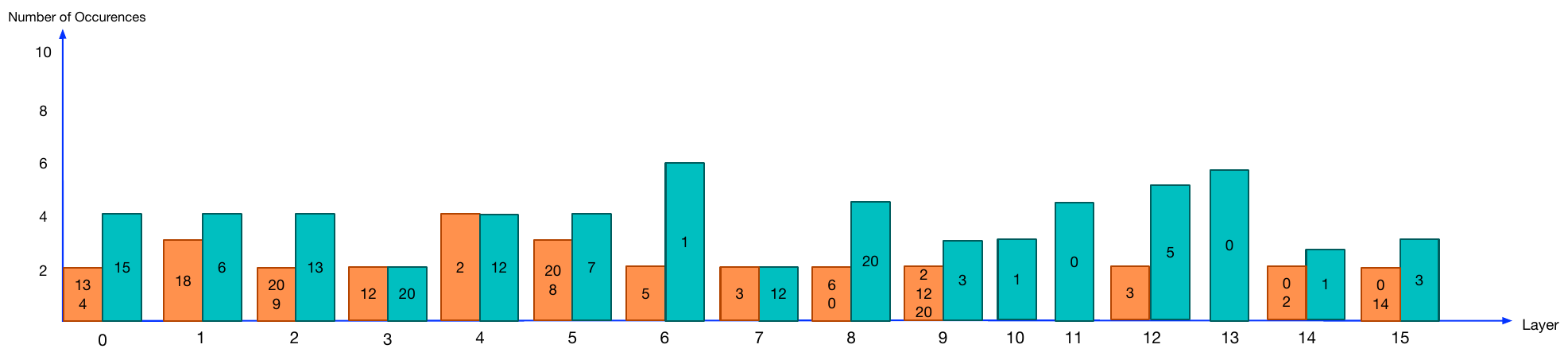}
    \caption{Block statistics for each layer of top 10 subnets on CIFAR-10. The horizontal axis represents the layer id of subnets. For instance, the "0" means that the first layer on the subnets. The vertical axis represents which two blocks are the most frequently \revised{adopted} in certain layer. For instance, block id 15 is the most frequently being used in the first layer of top 10 subnets. And block id 13 and block id 4 is the second most frequently being used in the first layer.}
    \label{fig:block_mode_statistics_cifar10}
\end{figure*}

\begin{figure*}[htb]
    \centering
    \includegraphics[width=0.9\textwidth]{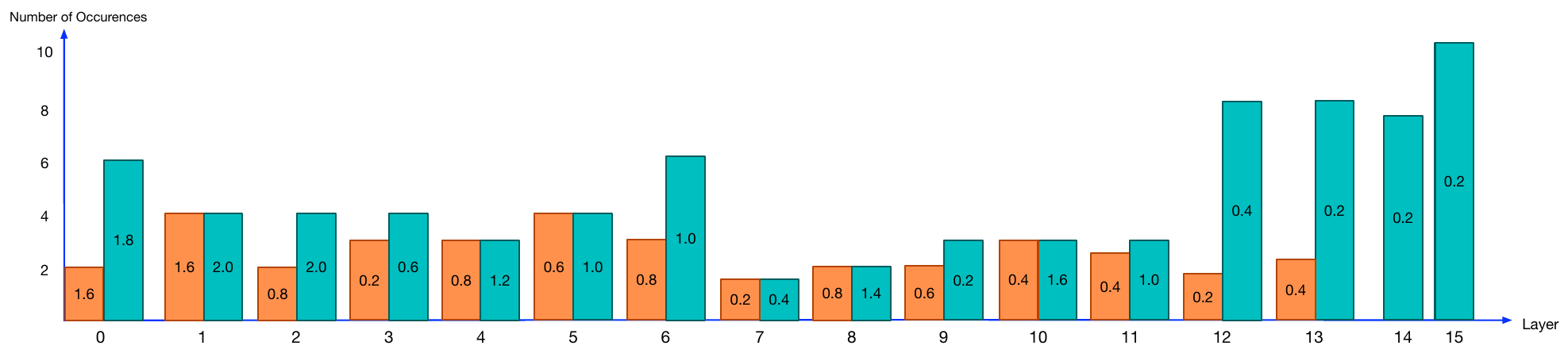}
    \caption{Channel statistics for each layer of top 10 subnets on CIFAR-10. The horizontal axis represents the layer id of subnets. The vertical axis represents which two channel expansion ratios are the most frequently being used in certain layer. For instance, channel expansion ratio 1.8 is the most frequently being used in the first layer of top 10 subnets and channel expansion ratio 1.6 is the second most frequently being used in the first layer.}
    \label{fig:channel_mode_statistics_cifar10}
\end{figure*}

\subsubsection{\textbf{Block and Channel Analysis}}
We make another statistics analysis of which block and channel expansion ratio are used most frequently in the top 10 subnets from the search results. Our assumption is that the best trade-off tiny neural network architectures can be viewed as a combinatorial optimization problem. Namely, each layer of network architecture could be viewed as the combination of certain blocks and channels and the optimum network architecture could be viewed as different layers of neural network combinatorial optimization problem.

Table. \ref{tab:block_encoding_statistics_cifar10} shows our block encoding scheme and layer statistics. The upper part of Table. \ref{tab:block_encoding_statistics_cifar10} explains how to build up the certain block for each block identifier. For instance, Block 0 is composed by ShuffleV2 block and SE layer. The kernel size of Block 0 is 3. The specific block internal composition can be referred in Fig. \ref{fig:shuffl2v2_block_choices} (a). In addition, Fig. \ref{fig:block_mode_statistics_cifar10} shows which two blocks are the most frequently adopted in a certain layer. The data in Fig. \ref{fig:block_mode_statistics_cifar10} comes from Table. \ref{tab:block_encoding_statistics_cifar10}. For our block encoding scheme, block identifies less than eight denote pure tiny blocks. Block identifiers between 9 and 17 are used to enhance the robustness of the tiny blocks. Block identifiers larger than 18 denote pure robust blocks. From Fig. \ref{fig:block_mode_statistics_cifar10}, we can observe the trend that the top trade-off tiny neural networks prefer to use robust blocks or tiny robust blocks in the first four layers while the last eight layers prefer to adopt pure tiny blocks. Our explanation is that since PGD attack mainly focuses on pixel-wise perturbations, the robust blocks are able to mitigate the attack effect in the first several layers and the tiny blocks can help the neural network to keep the balance between clear performance and the number of parameters. 

\begin{table*}[ht]
    \centering
    \setlength{\tabcolsep}{0.8mm}
    \caption{The Lego Subnet Experiments on CIFAR-10(\%)}
    \begin{tabular}{c|c|c|cccccccccccc}
    Supernet Training& Subnet Model Size & Clean Acc & $\textbf{P}_{2/255}^{10}$ & $\textbf{P}_{4/255}^{10}$ & $\textbf{P}_{6/255}^{10}$ & $\textbf{P}_{8/255}^{10}$ & $\textbf{P}_{2/255}^{30}$ & $\textbf{P}_{4/255}^{30}$ & $\textbf{P}_{6/255}^{30}$ & $\textbf{P}_{8/255}^{30}$ & $\textbf{P}_{2/255}^{50}$ & $\textbf{P}_{4/255}^{50}$ & $\textbf{P}_{6/255}^{50}$ & $\textbf{P}_{8/255}^{50}$\\
    \hline
    $\textbf{LEGO 2S}_{8/255}^{1}$ & \textbf{2.1546M} & \textbf{80.495} & \textbf{70.64} & \textbf{59.16} & \textbf{46.70} & \textbf{35.02} & \textbf{70.65} & \textbf{59.14} & \textbf{46.60} & \textbf{34.51} & \textbf{70.67} & \textbf{59.12} & \textbf{46.61} & \textbf{34.42} \\
    \hline
    $\textbf{2S}_{8/255}^{1}$ & 2.1878M & 79.87 & 69.38 & 56.81 & 43.61 & 31.73 & 69.38 & 56.71 & 43.51 & 31.12 & 69.37 & 56.76 & 43.44 & 31.09 \\ 
    \hline
    \end{tabular}\label{tab:lego_cifar10}
\end{table*}

\begin{table*}[ht]
    \centering
    \setlength{\tabcolsep}{0.8mm}
    \caption{The Lego Subnet Experiments on SVHN(\%)}
    \begin{tabular}{c|c|c|cccccccccccc}
    Supernet Training& Subnet Model Size & Clean Acc & $\textbf{P}_{2/255}^{10}$ & $\textbf{P}_{4/255}^{10}$ & $\textbf{P}_{6/255}^{10}$ & $\textbf{P}_{8/255}^{10}$ & $\textbf{P}_{2/255}^{30}$ & $\textbf{P}_{4/255}^{30}$ & $\textbf{P}_{6/255}^{30}$ & $\textbf{P}_{8/255}^{30}$ & $\textbf{P}_{2/255}^{50}$ & $\textbf{P}_{4/255}^{50}$ & $\textbf{P}_{6/255}^{50}$ & $\textbf{P}_{8/255}^{50}$\\
    \hline
    $\textbf{LEGO 2S}_{8/255}^{5}$ & \textbf{2.1593M} & \textbf{92.81} & \textbf{82.81} & \textbf{76.54} & \textbf{68.81} & \textbf{61.88} & \textbf{82.79} & \textbf{76.51} & \textbf{68.75}& \textbf{61.55} & \textbf{82.75} & \textbf{76.48} & \textbf{68.73} & \textbf{61.53} \\
    \hline
    $\textbf{2S}_{8/255}^{1}$ & 2.1676M & 91.87 & 79.80 & 73.45 & 65.31 & 57.43 & 79.71 & 73.42 & 65.27 & 57.38 & 79.70 & 73.40 & 65.18 & 57.32 \\ 
    \hline
    \end{tabular}\label{tab:lego_svhn}
\end{table*}

Fig.\ref{fig:channel_mode_statistics_cifar10} shows which two channel expansion ratios are used most frequently in certain layer. From Fig. \ref{fig:channel_mode_statistics_cifar10}, we can observe that the top trade-off tiny neural networks prefer to adopt larger channel in the first three layers and then the channel number for the rest of layers gradually declines. The top 10 tiny subnets use the smallest channel number in the last three layers. Our explanation is that since PGD attack mainly focuses on pixel-wise perturbations, the wider channel in the first several layers can help to mitigate the adversarial attacks, and the gradually declining channel number is able to maintain the tiny size of neural network.
\subsubsection{\textbf{Assumption Verification}}
In order to verify our assumption, the most frequently used blocks and channels of each layer in Fig. \ref{fig:block_mode_statistics_cifar10} and Fig. \ref{fig:channel_mode_statistics_cifar10} are selected as the block and channel choices, respectively, in our model. We call this new subnet architecture as Lego-Net. Table. \ref{tab:lego_cifar10} shows that Lego-Net performs better than our state-of-the-art subnet $2S^{1}_{8/255}$, especially in adversarial performance. Specifically, Lego-Net is able to increase adversarial accuracy by 10.36\% in $P_{8/255}^{10}$ and clean accuracy by 0.78\% while the size of Lego-Net drops by 1.52\%. We also achieve the same result on SVHN dataset shown in Table. \ref{tab:lego_svhn}. In conclusion, in order to build up tiny robust neural networks, we should put more pure robust or tiny robust blocks in the shallow layers and pure tiny blocks in the rest of layers. In term of channels design, we should put wider intermediate channels in the shallow layers and gradually reduce the intermediate channels in the rest of layers.

\begin{table}[th]
    \centering
    \setlength{\tabcolsep}{0.5mm}
    \caption{Benchmark on CIFAR-10(\%)}
    \begin{tabular}{c|c|c|cccccccccccc}
    Supernet Training& Subnet Model Size & Clean Acc & $\textbf{P}_{8/255}^{20}$ & $\textbf{P}_{8/255}^{100}$ & $\textbf{P}_{2/255}^{100}$\\
    \hline 
    \textbf{\revised{ResNet-18}} & 11.17M & 78.38 & 45.60 & 45.10 & 69.12 \\
    \hline
    \textbf{\revised{ShuffleNetV2}} & 4.82M & 75.23 & 25.69 & 25.12 & 58.12\\ 
    \hline
    \textbf{\revised{MobileNetV3}} & 4.93M & 76.13 & 26.73 & 26.71 & 60.34 \\ 
    \hline
    \textbf{\revised{RobNet-medium}}\cite{guo2020meets} & 5.66M & 78.33 & 49.13 & 48.96 & -- \\
    \hline
    $\textbf{RobNet-small}\cite{guo2020meets}$ & 4.41M & 78.05 & \textbf{48.32} & \textbf{48.07} & -- \\
    \hline
    $\textbf{LEGO 2S}_{8/255}^{1}$ & 2.1546M & \textbf{80.50} & 35.01 & 34.41 & \textbf{70.66}\\
    \hline
    $\textbf{2S}_{8/255}^{1}$ & 2.1878M & 79.87 &31.73 & 31.09 & 69.35\\ 
    \hline
    $\textbf{S}_{6/255}^{10}$ & 1.6840M & 78.63 & 29.02 & 28.07 & 67.75 \\
    \hline
    $\textbf{S}_{8/255}^{2}$ & \textbf{1.6822M} & 76.54 & 31.80 & 31.25 & 66.35 \\
    \hline
    \end{tabular}\label{tab:benchmark_cifar10}
\end{table}

\subsection{\textbf{Comparison}}
\revised{From Table \ref{tab:benchmark_cifar10} we can clearly see that the accuracy and the size of LEGO $2S_{8/255}^{1}$ outperforms the handcraft tiny neural network, such as ShuffleNetV2, MobileNetV3 and ResNet18. LEGO $2S_{8/255}^{1}$ also achieves a higher clean accuracy and more tiny size in comparison with RobNet-small \cite{guo2020meets}. Furthermore, we hypothesize that the epsilon size (the pixel perturbation range) in reality is not so much high as 8 pixels and the most common attack should be light-weight perturbation. So we try to reduce the epsilon size from $8/255$ to $2/255$ and we find that the gap between the adversarial performance and clean performance for our LegoNet is only 14\%, which is better than RobNet-small. Moreover, although RobNet-small is claimed to be one-shot, they use different computational budgets: small, medium and large to search different sizes of neural networks. In other words, they at least search three times to get different sizes of neural networks. Our work is a real one-shot NAS algorithm as the multi-objective optimization algorithm employed in our method is able to generate diverse models with different structures in one-shot}.

\section{Conclusion}
We propose a tiny adversarial multi-objective oneshot neural network search framework, which aims to find the best trade-off networks in terms of the adversarial error, the clean error and the size of neural network. Our study revealed several observations on how the adversarial training method of supernet will affect the subnets' adversarial performance. We also give a hint about how to design tiny robust neural networks based on our blocks statistics and channel statistics. We also conduct experiments to quantitatively prove our hints on improving the robustness of neural network without significantly reducing the clean accuracy and enlarging the size of neural network. However, there is still a drawback for our TAM-NAS framework. The performance of subnets largely rely on the network architecture of the supernets. So in the future, we may propose a co-evolutionary multi-objective NAS framework, i.e., the network architecture of supernet will also evolve during the search pharse.


\bibliographystyle{IEEEtran}
\bibliography{ref}

\end{document}